\documentclass{article}
\usepackage{amsmath}
\usepackage{lipsum}
\usepackage{hyperref}  
\usepackage{indentfirst}  
\numberwithin{equation}{section}  
\newtheorem{theorem}{Theorem}[section]  
\newtheorem{definition}{Definition}[section]  
\usepackage{graphicx}  

\numberwithin{table}{section}
\usepackage{subcaption} 
\usepackage{caption} 
\usepackage{xcolor}  
\usepackage{algorithm} 
\usepackage{algorithmic} 
\usepackage{natbib}  
\setcitestyle{authoryear,round}  
\usepackage{authblk}
\usepackage{comment}
\usepackage{tabularx}
\usepackage{booktabs}
\usepackage{multirow}

\providecommand{\keywords}[1]
{
	\textbf{\text{keywords:}} #1
	}

\title{Clustering based on Stochastic Dominance with application for risk averters and risk seekers}
\author[a]{Hua Li}
\author[b]{Xue Jia}
\author[a]{Yilin Kang}
\author[c]{Wing-Keung Wong}
\affil[a]{School of Science, Changchun University, Changchun, China.}
\affil[b]{School of Mathematics and Science, Northeast Normal University, Changchun, China.}

\date{}

\begin{document}
\begin{sloppypar}
	\maketitle 
	\begin{abstract}
		Stochastic Dominance (SD) theory provides a rigorous framework for selecting superior assets tailored to the asset allocation needs of investors with varying risk preferences (i.e., risk-averse, risk-seeking, and risk-neutral). However, traditional stock clustering methods typically rely on geometric metrics such as Euclidean distance, which often fail to effectively capture the intrinsic risk dominance relationships among assets. To address this limitation, this paper proposes an innovative clustering analysis framework based on SD test statistics. Methodologically, this study deeply integrates SD theory with machine learning algorithms. Transcending the limitations of traditional reliance on geometric distance, we innovatively utilize test statistics from first-, second-, and third-order SD to construct a "Stochastic Dominance Coefficient Matrix." Building upon this matrix, we modify the classic K-means and Hierarchical Clustering algorithms. Specifically, we derive 12 distinct algorithm variants tailored to different orders of SD relationships. Simultaneously, we construct the SD-SC coefficient and the SD-DBI index as specialized validity indices to evaluate the clustering performance. Empirically, we analyze constituent stock data from a representative developed market (the US NASDAQ Index) and an emerging market (China's CSI 100 Index). The results verify the effectiveness and robustness of the proposed method. Furthermore, we apply the clustering results to the modification of the Single Index Model and the construction of Global Minimum Variance Portfolios (GMVP). The findings demonstrate that the proposed method effectively facilitates customized asset allocation for investors, holding significant theoretical value and practical implications.

	\end{abstract}
	
	\section{Introduction}
	
	Stock clustering algorithms play a pivotal role in quantitative finance and the asset management industry, serving as a core mechanism for understanding market complexity and conducting asset preselection. Their intrinsic value lies in enabling investors to identify the true underlying structure of the stock market, thereby categorizing stocks with similar return characteristics or risk profiles into distinct groups. This data-driven market segmentation not only significantly reduces the computational dimensionality involved in portfolio construction but also provides a solid foundation for formulating differentiated investment strategies.
	
	A review of existing literature reveals that scholars both domestic and international have achieved fruitful results in stock clustering. Traditional clustering research predominantly employs classic machine learning algorithms: \cite{Yang2019} and \cite{Wu2022} utilized the K-means algorithm for stock partitioning; \cite{Huang2010} and \cite{Lu2020} explored the sectoral structures of the SSE 50 Index and other markets based on Agglomerative Hierarchical Clustering (AHC) and Spectral Clustering; \cite{Korzeniewski2018} further introduced the Partitioning Around Medoids (PAM) algorithm to construct portfolios with enhanced risk resistance. In recent years, with the advancement of deep learning, \cite{Lúcio2022} and \cite{Siregar2024} have attempted to incorporate time-series models (such as TGARCH) or specific market features (e.g., Indonesian stock data) into clustering frameworks. However, despite their respective merits in capturing market trends, these methods share a common limitation: traditional stock clustering approaches predominantly rely exclusively on stock-specific information (e.g., price, volatility, or financial metrics), neglecting the heterogeneity of market participants—namely, the "investors".
	
	In reality, investors are typically categorized into three distinct types based on their risk preferences: risk-averse, risk-seeking, and risk-neutral. Divergent risk attitudes inevitably lead to fundamentally different asset selection logic. In current literature, stock selection based on risk preferences often relies on expert experience or heuristic indicators. For instance, \cite{Guy2014} utilized decomposed Beta coefficients to assist risk decisions, while \cite{Boloș2025} defined "defensive assets" based on financial metrics such as ROA and Gross Profit Margin. Such approaches, while intuitive, often lack mathematical rigor and are prone to subjective judgment. To overcome this defect, Stochastic Dominance (SD) theory offers an ideal solution. In contrast to traditional Mean-Variance analysis, SD theory does not require the assumption that returns follow a normal distribution, nor does it restrict investors to quadratic utility functions. Thus, it provides a mathematically rigorous framework for selecting superior assets tailored to investors with varying preferences.
	
	In the development of SD theory, \cite{McFadden1989} first developed a test for risk-averse populations using minimum/maximum statistics. Scholars such as \cite{Klecan1991}, \cite{Barrett2003}, and \cite{Linton2005} subsequently refined this work by relaxing the Independent and Identically Distributed (IID) assumption. For risk-seeking populations, \cite{McAleer2006} and \cite{Sriboonchitta2009} proposed corresponding test methods, and \cite{Bai2015} integrated tests for both risk-averse and risk-seeking groups into a unified framework. Despite the extensive application of SD theory in finance (\cite{Levy1985}; \cite{Post2003}; \cite{Wong2007}), its traditional applications are mostly limited to pairwise asset comparisons or serving as constraints in optimization problems. Directly applying SD to stock selection within the Chinese stock market, which houses thousands of listed companies, would necessitate tens of thousands of pairwise tests, resulting in a prohibitive computational burden that is difficult to implement in practice.
	
	To address these pain points—specifically, how to efficiently identify asset classes that align with specific risk preferences (risk-averse or risk-seeking) within a large-scale market—this study innovatively proposes a clustering analysis method based on Stochastic Dominance test statistics. Distinct from traditional clustering that relies on Euclidean distance, we deeply integrate Stochastic Dominance theory with machine learning algorithms:
	
	\quad 1. First, we utilize the test statistics from first-, second-, and third-order Stochastic Dominance associated with different investor preferences to construct a core "Stochastic Dominance Coefficient Matrix."
	
	\quad 2. Second, building upon this matrix, we propose improved SD-K-means and SD-Hierarchical clustering algorithms. These new algorithms are capable of automatically categorizing stocks according to specific investor preferences, enabling investors to rapidly lock onto a desired pool of assets, thereby greatly simplifying the subsequent portfolio construction process.
	
	\quad 3. Furthermore, this paper introduces novel clustering validity indices (SD-SC coefficient and SD-DBI index) specifically designed for these new models, further refining the theoretical framework of preference-based stock clustering.
	
	The remainder of this paper is organized as follows: Section 2 introduces the relevant theoretical background, including Stochastic Dominance tests and the Bootstrap method. Section 3 details the construction process of the proposed clustering algorithms based on SD theory, as well as the novel indices used for evaluating clustering performance. Section 4 demonstrates the performance of the proposed method through empirical research based on real-world market data. Section 5 concludes the paper and discusses the practical implications of this method for customized asset allocation as well as directions for future research.
	
	\section{SD Test and Bootstrap Method} \label{sec:SD Test and Bootstrap Method}
	\subsection{ASD(DSD) theorem}
	we let $X$ and $Y$ be random variables defined on $[a, b]$ with cumulative distribution functions (CDFs) $F$ and $G$, and probability density functions (PDFs), $f$ and $g$, respectively. We define \begin{equation}\label{equ:1}
		\begin{split}
			&h(x) = H_{0}^{A}(x) = H_{0}^{D}(x) \\
			&H_{j}^{A}(x) = \int_{a}^{x} H_{j-1}^{A}(y) dy, j=2,3\\
			&H_{j}^{D}(x) = \int_{x}^{b} H_{j-1}^{D}(y) dy, j=2,3
		\end{split} 
	\end{equation} 
	where $H=F$ or $G$ and $h = f$ or $g$. $\mu_{F}=\mu_{X}=\int_{a}^{b}x d F(x)$ is the mean of $X$ and $\mu_{G}=\mu_{Y}=\int_{a}^{b}x d G(x)$ is the mean of $Y$.
	
	We note that $H_{j}^{A}$ in (\ref{equ:1}) can be used to develop the SD theory for risk averters and thus we call this type of SD the ascending SD (ASD) and call $H_{j}^{A}$ the $j$th order ASD integral or the $j$th order cumulative probability, since $H_{j}^{A}$ is integrated from $H_{j-1}^{A}$ in ascending order from the leftmost point of downside risk. On the other hand, $H_{j}^{D}$ in (\ref{equ:1}) can be used to develop the SD theory for risk seekers and thus we call this type of SD the descending SD (DSD) and call $H_{j}^{D}$ the $j$th order DSD integral or the $j$th order reversed cumulative probability, since $H_{j}^{D}$ is integrated from $H_{j-1}^{D}$ in descending order from the rightmost point of upside profit. so we give the defination as follows \cite{Li2013}:
	
	\begin{definition}
		Give $X$ and $Y$ be random variables with cumulative distribution functions (CDFs) $F$ and $G$, in the sense of FASD (SASD, TASD), $X$ dominance $Y$ or $F$ dominance $G$, denote $X\succeq_1^AY$ or $F\succeq_1^AG$ ($X\succeq_2^AY$ or $F\succeq_2^AG$, $X\succeq_{3}^{A}Y$ or $F\succeq_{3}^{A}G$), if and only if, any $x\in[a,b]$, $F_1^A(x)\leq G_1^A(x) (F_2^A(x)\leq G_2^A(x)$, $F_3^A(x)\leq G_3^A(x))$, where FASD, SASD, TASD respectively represet First-order, Second-order, Third-order ascending stochastic dominance.
	\end{definition}
	
	\begin{definition}
		Give $X$ and $Y$ be random variables with cumulative distribution functions (CDFs) $F$ and $G$, in the sense of FDSD (SDSD, TDSD), $X$ dominance $Y$ or $F$ dominance $G$, denote $X\succeq_1^DY$ or $F\succeq_1^DG$ ($X\succeq_2^DY$ or $F\succeq_2^DG$, $X\succeq_{3}^{D}Y$ or $F\succeq_{3}^{D}G$), if and only if, any $x\in[a,b]$, $F_1^D(x)\geq G_1^D(x) (F_2^D(x)\geq G_2^D(x)$, $F_3^D(x)\geq G_3^D(x))$, where FDSD, SDSD, TDSD respectively represet First-order, Second-order, Third-order descending stochastic dominance.
	\end{definition}
	
	The ASD and DSD approaches are regarded as two of the most useful tools for ranking uncertain investment prospects, \citet{Li1999} has shown that ranking assets is equivalent to utility maximization for both risk averters and risk seekers. The existence of ASD implies that the expected utility of the risk-averse investor is always higher when holding the dominant asset than when holding the dominated asset, and consequently, the dominated asset would not be chosen. The same is true for DSD \citep{Bai2015}. 
	
	\subsection{Stochastic Dominance Test}
	According to the definition of ASD and DSD, we can test the following null hypothesis for $j=1, 2, 3$:
	\begin{equation}\label{equ:2}
		H_{0}^{A(D)}: F_{j}^{A(D)}\equiv G_{j}^{A(D)}
	\end{equation}
	
	Three opposing hypotheses:
	\begin{equation}\label{equ:3}
		\begin{split}
			H_{1}^{A(D)}: 	F\not \equiv _{j}^{A(D)}G\\ 
			H_{1l}^{A(D)}: 	F\succ_{j}^{A(D)}G \\
			H_{1r}^{A(D)}: 	F\prec _{j}^{A(D)}G 
		\end{split}
	\end{equation}
	
	Now assume that $\{f_{i};i=1,2,\cdots,N_{f}\}$ and $\{g_{i};i=1,2,\cdots,N_{g}\}$ are observations from independent random variables $X$ and $Y$, respectively, with distribution functions $F$ and $G$. The integrals $F_{j}^{A(D)}$ and $ G_{j}^{A(D)}$ of $F$ and $G$ are given by (\ref{equ:1}), $j=1,2,3$. For a given set of grid points $\{x_k,k=1,\cdots,K\}$, \citet{Davidson2000} propose to test $H_{1}^{A}$, $H_{1l}^{A}$ and $H_{1r}^{A}$ using the following $j$-th ASD test statistic, $T_j^A(x)$, and \citet{Wong2007} propose to test $H_{1}^{D}$, $H_{1l}^{D}$ and $H_{1r}^{D}$ using the following $j$-th DSD test statistic, $T_j^D(x)$:
	
	\begin{equation}\label{equ:4}
		T_j^A(x)=\frac{\hat{F}_j^A(x)-\hat{G}_j^A(x)}{\sqrt{\hat{V}_j^A(x)}}
	\end{equation}
	
	\begin{equation}\label{equ:5}
		T_j^D(x)=\frac{\hat{F}_j^D(x)-\hat{G}_j^D(x)}{\sqrt{\hat{V}_j^D(x)}}
	\end{equation}
	Here,
	\begin{equation}\label{equ:6}
		\begin{array}{rcl}{{\hat{V}_{j}^{A}(x)}}&{=}&{{\hat{V}_{F_{j}}^{A}(x)+\hat{V}_{G_{j}}^{A}(x)}}\\{{\hat{H}_{j}^{A}(x)}}&{=}&{{\frac{1}{N_{h}(j-1)!}\sum_{i=1}^{N_{h}}(x-h_{i})_{+}^{j-1}}}\\{{\hat{V}_{H_{j}}^{A}(x)}}&{=}&{{\frac{1}{N_{h}}\left[\frac{1}{N_{h}((j-1)!)^{2}}\sum_{i=1}^{N_{h}}(x-h_{i})_{+}^{2(j-1)}-\hat{H}_{j}^{A}(x)^{2}\right],H=F,G;h=f,g;}}\end{array}
	\end{equation}
	\begin{equation}\label{equ:7}
		\begin{array}{rcl}{{\hat{V}_{j}^{D}(x)}}&{=}&{{\hat{V}_{F_{j}}^{D}(x)+\hat{V}_{G_{j}}^{D}(x)}}\\{{\hat{H}_{j}^{D}(x)}}&{=}&{{\frac{1}{N_{h}(j-1)!}\sum_{i=1}^{N_{h}}(h_{i}-x)_{+}^{j-1}}}\\{{\hat{V}_{H_{j}}^{D}(x)}}&{=}&{{\frac{1}{N_{h}}\left[\frac{1}{N_{h}((j-1)!)^{2}}\sum_{i=1}^{N_{h}}(h_{i}-x)_{+}^{2(j-1)}-\hat{H}_{j}^{D}(x)^{2}\right],H=F,G;h=f,g;}}\end{array}
	\end{equation}
	
	Under the null hypothesis, \cite{Bai2015} using the limiting distribution of stochastic processes derives some theorems. 
	\begin{theorem}\citep[Theorem 3.1]{Bai2015}\label{th:1}
		Let $\{f_{i}\}(i=1,2,\cdots,N_{f})$ and $\{g_{i}\}(i=1,2,\cdots,N_{g})$ be random observations drawn from the independent random variables $X$ and $Y$, with CDFs, $F$ and $G$, respectively. Under the null hypothesis $F\equiv G$, the $j$-order
		ASD[DSD] test statistics, $T_j^A(x)[T_j^D(x)](j=1$, $2$, and $3)$, weakly tends to a limiting Gaussian process with mean 0,
		variance 1, and correlation function $r^A(x,y)[r^D(x,y)]$ in which for the case $j=1$, we get
		\[
		r_1^A(x,y)=\frac{F(x\wedge y)-F(x)F(y)}{\sqrt{F(x)F(y)(1-F(x))(1-F(y))}}
		\]
		\[
		r_1^D(x,y)=\frac{F(x\wedge y)-F(x)F(y)}{\sqrt{F(x)F(y)(1-F(x))(1-F(y))}}
		\]
		and for the case $j>1$, we have
		\[
		r^A(x,y)=\frac{\int_{a}^{x}\int_{a}^{y}(x-t)^{j-2}(y-s)^{j-2}(F(t\wedge s)-F(t)F(s))dtds}{\sqrt{V_j^A(x)V_j^A(y)}}
		\]
		\[
		r^D(x,y)=\frac{\int_{x}^{b}\int_{y}^{b}(t-x)^{j-2}(s-y)^{j-2}(F(t\wedge s)-F(t)F(s))dtds}{\sqrt{V_j^D(x)V_j^D(y)}}
		\]
		where $\hat{H}_j^A(x)$ and $\hat{V}_j^A(x)$ are defined in (\ref{equ:6}) and $\hat{H}_j^D(x)$ and $\hat{V}_j^D(x)$ are defined in (\ref{equ:7}) for $H=F$ and $G$, $V_1^A(x)=F(x)(1-F(x))$, $V_j^A(x)=\int_{a}^{x}\int_{a}^{x}(x-t)^{j-2}(x-s)^{j-2}(F(t \wedge s)-F(t)F(s))dtds$, $V_1^D(x)=F(x)(1-F(x))$, and  $V_j^D(x)=\int_{x}^{b}\int_{x}^{b}(t-x)^{j-2}(s-x)^{j-2}(F(t \wedge s)-F(t)F(s))dtds$.
	\end{theorem}

	Based on this theorem, rejection regions for the aforementioned hypothesis tests were proposed as follows.
	
	We reject the null hypothesis $H_{0}^{A}$ if:
	\begin{equation}
		\begin{split}
			&\mathop{max}\limits_{a\textless x \textless b}|T_{j}^{A}(x)| \textgreater M_{\infty,\alpha/2}^{A},\ \text{if the alternative hypothesis is $H_{1}^{A}$};\\
			&\mathop{min}\limits_{a\textless x \textless b}T_{j}^{A}(x) \textless -M_{\infty,\alpha}^{A},\ \text{if the alternative hypothesis is $H_{1l}^{A}$};\\
			&\mathop{max}\limits_{a\textless x \textless b}T_{j}^{A}(x) \textgreater M_{\infty,\alpha}^{A},\ \text{if the alternative hypothesis is $H_{1r}^{A}$};
		\end{split}
	\end{equation}
	
	We reject the null hypothesis $H_{0}^{D}$ if:
	\begin{equation}
		\begin{split}
			&\mathop{max}\limits_{a\textless x \textless b}|T_{j}^{D}(x)| \textgreater M_{\infty,\alpha/2}^{D},\ \text{if the alternative hypothesis is $H_{1}^{D}$};\\
			&\mathop{max}\limits_{a\textless x \textless b}T_{j}^{D}(x) \textgreater M_{\infty,\alpha}^{D},\ \text{if the alternative hypothesis is $H_{1l}^{D}$};\\
			&\mathop{min}\limits_{a\textless x \textless b}T_{j}^{D}(x) \textless -M_{\infty,\alpha}^{D},\ \text{if the alternative hypothesis is $H_{1r}^{D}$};
		\end{split}
	\end{equation}
	Here the critical value $M_{\infty,\alpha}^{A}$ and $M_{\infty,\alpha}^{D}$ are given by the Bootstrap method, the exact procedure will be described in detail later.
	
	\subsection{Bootstrap Methods and Critical Values}
	Assuming the returns of stocks are independent, and $\{f_{i};i=1,2,\cdots,N_{f}\}$ and $\{g_{i};i=1,2,\cdots,N_{g}\}$ are independently and identically distributed sample sequences, respectively. We put the two sample sets into one set $\{f_{(i)},g_{(j)},i=1,2,\cdots,N_{f},j=1,2,\cdots,N_{g}\}$ and perform sampling with replacement to obtain two resampled sample sets $\{f_{i}^{*},i=1,2,\cdots,N_{f}\}$ and $\{g_{i}^{*},i=1,2,\cdots,N_{g}\}$. In this way, Bootstrap calculation can be performed for different SD test statistics ($\hat{T}_{j}^{A}$, $\hat{T}_{j}^{D}$, $j=1,2,3$). By using this method, we can approximate the null distribution of the test statistic and obtain $\mathop{max}\limits_{a<x<b}|T_{j}^{A(D)}|,j=1,2,3$. The specific details are as follows:
	
	Step 1: Sample with replacement from the sample set $\{f_{(i)},g_{(j)},i=1,2,\cdots,N_{f},j=1,2,\cdots,N_{g}\}$ to obtain sample $\{f_{i}^{*},i=1,2,\cdots,N_{f}\}$, and do the same to get $\{g_{i}^{*},i=1,2,\cdots,N_{g}\}$.
	
	Step 2: Calculate
	\begin{equation}
		A_j=\max_{A<x<B}\left|\hat{T}_j^{*A}(x)\right|=\max_{A<x<B}\left|\frac{\hat{F}_j^{*A}(x)-\hat{G}_j^{*A}(x)}{\sqrt{\hat{V}_j^{*A}(x)}}\right|
	\end{equation}
	\begin{equation}
		D_j=\max_{A<x<B}\left|\hat{T}_j^{*D}(x)\right|=\max_{A<x<B}\left|\frac{\hat{F}_j^{*D}(x)-\hat{G}_j^{*D}(x)}{\sqrt{\hat{V}_j^{*D}(x)}}\right|
	\end{equation}
	Repeat the above steps $N$ times to obtain $N$ sets of $A_{j}$ and $D_{j}$, denoted as:
	
	\[
	\{A_{jk},D_{jk}, k=1,2,\cdots,N\}
	\]
	
	Step 3: Find $A_j(\alpha)$ such that $\#\{\left|A_{jk}\right| \ge A_j(\alpha),k\leq N\}=[N\alpha ]$, then $A_j(\alpha)$ is the $\alpha$ percentile of the corresponding distribution of $\hat{T}_j^{*A}(x)$.
	
	Step 4: Find $D_j(\alpha)$ such that $\#\{\left|D_{jk}\right| \ge D_j(\alpha),k\leq N\}=[N\alpha ]$, then $D_j(\alpha)$ is the $\alpha$ percentile of the corresponding distribution of $\hat{T}_j^{*D}(x)$.
	
	The preceding method makes a decision by comparing the test statistic with the critical value determined at a given significance level. While this decision rule provides a clear boundary, it only yields a binary conclusion of "reject" or "do not reject." Consequently, it fails to precisely characterize the degree of risk associated with the decision (i.e., the strength of the evidence). Therefore, to provide a more intuitive and precise reflection of statistical significance, this paper adopts the $p$-value as the ultimate criterion for decision-making.
	
	Specifically, let $T_{j,0}^A$ and $T_{j,0}^D$ denote the maximum observed values of the test statistics calculated from the original sample (i.e., $T_{j,0}^A = \max_{A<x<B} |\hat{T}_j^A(x)|$, $T_{j,0}^D = \max_{A<x<B} |\hat{T}_j^D(x)|$). Furthermore, let $\{A_{jk}, k = 1, 2, \cdots, N\}$ and $\{D_{jk}, k = 1, 2, \cdots, N\}$ represent the $N$ resampled test statistics obtained via the aforementioned Bootstrap procedure. The empirical $p$-value can then be calculated as follows:
	
	\begin{equation}
		p_j^A = \frac{\#\{A_{jk} \ge T_{j,0}^A, k = 1, 2, \cdots, N\}}{N}
	\end{equation}
	\begin{equation}
		p_j^D = \frac{\#\{D_{jk} \ge T_{j,0}^D, k = 1, 2, \cdots, N\}}{N}
	\end{equation}
	
	If the null hypothesis holds (i.e., there is no stochastic dominance relationship between the two assets), this $p$-value is typically large. Conversely, if the calculated $p$-value is sufficiently small, we are inclined to reject the null hypothesis, thereby concluding the existence of a stochastic dominance relationship between the two assets. Furthermore, the magnitude of the $p$-value can, to some extent, be viewed as a "distance" metric quantifying the deviation of the empirical distribution from the null hypothesis. A detailed discussion from this perspective will be elaborated in the subsequent chapter.
	
	\section{Clustering Algorithms Based on SD Test Statistics}
	
	In clustering algorithms, distance serves as a metric for the degree of dissimilarity between samples. To construct stock clustering models tailored to investors with different preferences, this paper calculates a Stochastic Dominance Coefficient Matrix using various stochastic dominance (SD) test statistics to measure the distance (or degree of dissimilarity) between stock clusters. Consequently, we propose the ASD-K-means, DSD-K-means, ASD-Hierarchical, and DSD-Hierarchical clustering algorithms. Based on the different orders of the Stochastic Dominance Coefficient Matrix, these can be further categorized into 12 specific algorithms: FASD-K-means, SASD-K-means, TASD-K-means, FASD-Hierarchical, SASD-Hierarchical, TASD-Hierarchical, FDSD-K-means, SDSD-K-means, TDSD-K-means, FDSD-Hierarchical, SDSD-Hierarchical, and TDSD-Hierarchical. Furthermore, the Bootstrap method is employed to estimate the empirical $p$-values required for the calculation of the Stochastic Dominance Coefficient Matrix.
	
	\subsection{Stochastic Dominance Coefficient}
	
	In the previous section, we introduced the test statistics $\hat{T}_j^A(x)$ and $\hat{T}_j^D(x)$ ($j=1,2,3$) for risk-averse (ASD) and risk-seeking (DSD) investors, respectively. Under the null hypothesis $F \equiv G$, the $j$-th order test statistics converge weakly to a limit Gaussian process. Using the Bootstrap method, we simulate the empirical distribution under the null hypothesis and calculate the empirical $p$-value as the ultimate criterion for decision-making. As is well known, if the null hypothesis holds (i.e., there is no stochastic dominance relationship between the two assets), this $p$-value is typically large. Conversely, if the calculated $p$-value is sufficiently small, we are inclined to reject the null hypothesis, thereby concluding the existence of a stochastic dominance relationship between the two assets.
	
	In traditional clustering analysis, algorithms inherently rely on the construction of a reasonable distance or similarity matrix. To seamlessly integrate the results of the stochastic dominance tests into a clustering framework, we transform the aforementioned $p$-values to construct the "Stochastic Dominance Coefficient." This coefficient is designed to satisfy the fundamental conditions of a distance metric: a larger coefficient indicates a more significant stochastic dominance relationship between $X$ and $Y$ (implying greater distinctness in their distributions). Conversely, a smaller coefficient suggests an insignificant stochastic dominance relationship (implying higher distributional similarity), meaning $X$ and $Y$ can be grouped into the same cluster. Specifically, we define the Stochastic Dominance Coefficient as the complement of the empirical $p$-value (i.e., $1 - p$). 
	
	In this paper, we propose novel clustering algorithms by employing the first-, second-, and third-order Stochastic Dominance Coefficients as distance measures between stocks. The specific calculation formulas for these coefficients are presented as follows:
	
	\begin{equation}
		P(X,Y)_j^A = 1 - \frac{\sum_{k=1}^N I(A_{jk} \ge T_{j,0}^A)}{N}, \quad j = 1, 2, 3
	\end{equation}
	
	\begin{equation}
		P(X,Y)_j^D = 1 - \frac{\sum_{k=1}^N I(D_{jk} \ge T_{j,0}^D)}{N}, \quad j = 1, 2, 3
	\end{equation}
	where $P(X,Y)_j^A$ denotes the Stochastic Dominance Coefficient corresponding to the risk-averse population; $X$ and $Y$ represent any arbitrary pair of stocks from the sample; $P(X,Y)_j^D$ denotes the Stochastic Dominance Coefficient corresponding to the risk-seeking population; and $j = 1,2,3$ correspond to the first-, second-, and third-order stochastic dominance coefficients, respectively. $I(\cdot)$ is the indicator function. Consistent with previous sections, $T_{j,0}^A$ and $T_{j,0}^D$ represent the maximum observed values of the test statistics for the risk-averse and risk-seeking populations, respectively, computed based on the original sample of log-returns for stock $X$ and stock $Y$. For the detailed calculation procedures of $A_{jk}$ and $D_{jk}$, please refer to Section 2.3.

	\subsection{Selection of the Number of Clusters $K$}
	In clustering algorithms, the selection of the number of clusters ($K$) is of paramount importance, as it directly impacts the quality of the model and the accuracy of the clustering results. The Silhouette Coefficient measures both intra-cluster compactness and inter-cluster separation; therefore, it is commonly employed to determine the optimal value of $K$. In this paper, by integrating the traditional Silhouette Coefficient with the specific characteristics of our proposed model, we construct a novel metric named the SD-SC Coefficient to determine the optimal number of clusters.
	
	The SD-SC Coefficient serves as a metric to evaluate the performance of the proposed stochastic dominance clustering method. It incorporates two components: cohesion and separation. Cohesion reflects the degree of compactness between a specific sample point and other sample points within the same cluster. For a given sample point $i$, cohesion is defined as the average stochastic dominance distance between that sample and all other samples within its assigned cluster. A smaller cohesion value indicates higher similarity among samples within the cluster and greater internal consistency, thereby implying superior clustering performance.
	
	Separation reflects the relationship between a sample point and elements outside its cluster. For a given sample point $i$, separation is defined as the average stochastic dominance distance between that sample and all sample points in the nearest neighboring cluster. A larger separation value implies significant differences between samples in distinct clusters and high discriminative power of the clustering results, indicating superior clustering performance.
	
	When applied to the stochastic dominance clustering model, the SD-SC Coefficient can scientifically assess the algorithm's effectiveness and determine the optimal number of clusters. In this study, the optimal number of clusters is jointly determined by the designated stochastic dominance clustering model and the corresponding SD-SC Coefficient. The specific procedures are as follows:
	
	Step 1: Define the search range for the number of clusters, denoted as $K \in [K_{\min}, K_{\max}]$.
	
	Step 2: Select a specific stochastic dominance clustering algorithm (e.g., ASD-K-means or DSD-Hierarchical) tailored to the target investor's risk preference.
	
	Step 3: For each candidate value $K$ within the defined search range, execute the selected clustering algorithm to obtain the corresponding clustering result.
	
	Step 4: Calculate the overall SD-SC Coefficient for each clustering result obtained in Step 3. The value of $K$ that yields the maximum SD-SC Coefficient is identified as the optimal number of clusters.
	
	The specific calculation procedures for the SD-SC Coefficient are detailed in Section 3.5.
	
	\subsection{ASD(DSD)-K-means Clustering Algorithm}
	First, we define the distance metric to measure the dissimilarity between stocks tailored to specific investor risk preferences: the Stochastic Dominance Coefficient. For any two arbitrary stocks $X$ and $Y$, the calculation methods for the stochastic dominance coefficients for risk-averse and risk-seeking investors are precisely defined in Equations (3.1) and (3.2), respectively. Consistent with the definitions in Section 3.1, a smaller coefficient indicates a higher degree of distributional similarity between the two assets.
	
	Subsequently, utilizing the Stochastic Dominance Coefficient Matrix as the distance metric, stocks with higher similarity can be grouped into the same cluster via the modified K-means algorithm, denoted as the ASD-K-means or DSD-K-means algorithm. Specifically, substituting the first-order ($j=1$), second-order ($j=2$), and third-order ($j=3$) stochastic dominance coefficients yields the FASD(FDSD)-K-means, SASD(SDSD)-K-means, and TASD(TDSD)-K-means clustering algorithms, respectively.
	
	In the standard K-means algorithm, the number of clusters $K$ must be designated in advance (as discussed in Section 3.2). A critical distinction in our proposed ASD(DSD)-K-means algorithm lies in the update mechanism of the cluster centers. During each iteration, the new cluster center is updated by calculating the cross-sectional average of the return series for all stocks currently assigned to that cluster, essentially forming an equally-weighted portfolio of the constituents. The specific execution process of the ASD(DSD)-K-means clustering algorithm is presented in Algorithm 1:
	
	\begin{algorithm}[H]
		\caption{ASD(DSD)-K-means Clustering Algorithm}
		\textbf{Input:} \\
		The return dataset of $n$ stocks, denoted as $R$; \\
		The designated number of clusters, $K$; \\
		The maximum number of iterations, $max\_iter = 100$; \\
		\textbf{Initialization:} \\
		Randomly select $K$ stocks from the sample as the initial cluster centers $\{C_t : t = 1, 2, \cdots, K\}$; \\
		Set iteration counter $ite = 0$;
		\begin{algorithmic}[1]
			\REPEAT
			\FOR{each stock $i = 1, 2, \cdots, n$}
			\FOR{each cluster center $t = 1, 2, \cdots, K$}
			\STATE Calculate the stochastic dominance distance $d_{i,t}$ between stock $R_i$ and cluster center $C_t$ using Equation (3.1) for ASD or Equation (3.2) for DSD.
			\ENDFOR
			\STATE Assign stock $i$ to the nearest cluster: $cluster\_idx_i \leftarrow \arg\min_t \{d_{i,t} : t = 1, 2, \dots, K\}$
			\ENDFOR
			
			\FOR{each cluster $t = 1, 2, \cdots, K$}
			\STATE Update the cluster center $C_t$ by calculating the mean return of all stocks currently assigned to cluster $t$: 
			$$ C_t = \text{mean}(R_{cluster\_idx == t}) $$
			\ENDFOR
			\STATE $ite \leftarrow ite + 1$
			\UNTIL{no stocks change their cluster assignments OR $ite == max\_iter$}
		\end{algorithmic}
		\textbf{Output:} The final cluster assignments for all $n$ stocks.
	\end{algorithm}
	
	\subsection{ASD(DSD)-Hierarchical Clustering Algorithm}
	Based on the order of hierarchical decomposition, hierarchical clustering algorithms can generally be categorized into Agglomerative (bottom-up) and Divisive (top-down) approaches. In this paper, we explicitly adopt the Agglomerative Hierarchical Clustering approach to construct our model. This algorithm proceeds in a bottom-up manner, initializing by treating each individual stock as an independent, single-element cluster. At each iterative step, it merges the two closest clusters together until all stocks are eventually agglomerated into one overarching cluster.
	
	During the merging process, determining the exact "distance" between two clusters is crucial. While standard clustering algorithms frequently rely on Euclidean or Mahalanobis distances, our proposed SD-Hierarchical clustering model innovatively adopts the Stochastic Dominance Coefficient Matrix as the fundamental distance metric. Furthermore, among the common inter-cluster similarity measurement strategies—namely Single Linkage (minimum distance), Complete Linkage (maximum distance), and Average Linkage—we select the Average Linkage method. This method comprehensively incorporates the structural information of all samples within the clusters, effectively mitigating the outlier sensitivity commonly associated with Single or Complete Linkage.
	
	Specifically, the Average Linkage method defines the distance between cluster $C_1$ and cluster $C_2$ as the arithmetic mean of the stochastic dominance distances between all possible pairs of stocks across the two clusters. The inter-cluster distances tailored to risk-averse and risk-seeking investors are mathematically defined as follows:
	
	\begin{equation}
		dist(C_1, C_2)_j^A = \frac{1}{|C_1| \cdot |C_2|} \sum_{x \in C_1} \sum_{y \in C_2} P(x, y)_j^A
	\end{equation}
	
	\begin{equation}
		dist(C_1, C_2)_j^D = \frac{1}{|C_1| \cdot |C_2|} \sum_{x \in C_1} \sum_{y \in C_2} P(x, y)_j^D
	\end{equation}
	where $|C_1|$ and $|C_2|$ represent the total number of stocks contained in clusters $C_1$ and $C_2$, respectively; $x$ and $y$ represent arbitrary individual stocks belonging to $C_1$ and $C_2$; and $P(x, y)_j^A$ and $P(x, y)_j^D$ are the pairwise Stochastic Dominance Coefficients calculated previously.
	
	By integrating the Agglomerative hierarchical framework with the SD Coefficient Matrix, we finalize the Stochastic Dominance Hierarchical clustering model. Specifically, substituting $j = 1, 2, \text{and } 3$ into the distance metrics denotes the FASD(FDSD)-, SASD(SDSD)-, and TASD(TDSD)-Hierarchical clustering algorithms, respectively. The specific step-by-step execution of the ASD(DSD)-Hierarchical clustering algorithm is presented in Algorithm 2:
	
	\begin{algorithm}[H]
		\caption{ASD(DSD)-Hierarchical Clustering Algorithm}
		\textbf{Input:} \\
		The return dataset of $n$ stocks, denoted as $R = \{R_1, R_2, \cdots, R_n\}$; \\
		The designated number of clusters, $K$; \\
		The specific order of stochastic dominance, $j \in \{1, 2, 3\}$. \\
		\textbf{Initialization:} \\
		Assign each stock to its own individual cluster: $C_i = \{R_i\}$ for $i = 1, 2, \cdots, n$; \\
		Define the initial active cluster set: $\mathcal{C} = \{C_1, C_2, \cdots, C_n\}$; \\
		Calculate the initial $n \times n$ pairwise distance matrix $M$, where $M_{u,v} = P(R_u, R_v)_j^A$ (or $P(R_u, R_v)_j^D$) using Equation (3.1) or (3.2) for all stocks $u, v \in \{1, 2, \cdots, n\}$.
		\begin{algorithmic}[1]
			\WHILE{the total number of clusters $|\mathcal{C}| > K$}
			\STATE Find the pair of clusters $C_a$ and $C_b$ in $\mathcal{C}$ that have the minimum inter-cluster distance $\text{dist}(C_a, C_b)_j^A$ (or $\text{dist}(C_a, C_b)_j^D$) based on the Average Linkage method calculated via Equation (3.3) or (3.4).
			\STATE Merge the two closest clusters into a new single cluster: $C_{new} = C_a \cup C_b$.
			\STATE Update the active cluster set: remove $C_a$ and $C_b$, and insert $C_{new}$ ($\mathcal{C} \leftarrow \mathcal{C} \setminus \{C_a, C_b\} \cup \{C_{new}\}$).
			\STATE Update the distance matrix $M$ by calculating the new average linkage distances between $C_{new}$ and all other remaining clusters in $\mathcal{C}$.
			\ENDWHILE
		\end{algorithmic}
		\textbf{Output:} The final $K$ clusters and the names of the stocks contained within each class.
	\end{algorithm}
	
	\subsection{SD-SC Coefficient}
	The Silhouette Coefficient (SC) is a classic clustering validity index used to measure the compactness and separation of clustering results. Traditional SC calculations rely on Euclidean or Manhattan distances; however, these metrics fail to account for investors' risk preferences and thus cannot adequately reflect the clustering performance of the model proposed in this paper. To address this limitation, this subsection proposes a novel evaluation metric based on the stochastic dominance distance, termed the SD-SC Coefficient, to better assess the clustering quality.
	
	The SD-SC Coefficient is an enhanced clustering validity index derived from the traditional Silhouette Coefficient by substituting the standard distance metric with the stochastic dominance distance. The calculation steps are as follows:
	
	Step 1: Calculate the pairwise stochastic dominance distances between all stocks. For any two arbitrary stocks $i$ and $v$, their distances corresponding to risk-averse and risk-seeking investors are denoted as $P(i, v)_j^A$ and $P(i, v)_j^D$, respectively.
	
	Step 2: Calculate the intra-cluster distance $a(i)_j$: For a given sample point $i$ assigned to cluster $C_i$, calculate the average stochastic dominance distance between $i$ and all other points within the same cluster $C_i$.
	
	\begin{equation}
		a(i)_j^A = \frac{1}{|C_i| - 1} \sum_{v \in C_i, v \neq i} P(i, v)_j^A
	\end{equation}
	
	\begin{equation}
		a(i)_j^D = \frac{1}{|C_i| - 1} \sum_{v \in C_i, v \neq i} P(i, v)_j^D
	\end{equation}
	where $|C_i|$ denotes the total number of samples in cluster $C_i$; $a(i)_j^A$ and $a(i)_j^D$ represent the intra-cluster distances corresponding to risk-averse and risk-seeking investors under the $j$-th order SD, respectively.
	
	Step 3: Calculate the inter-cluster distance $b(i)_j$: For sample point $i$, calculate the average stochastic dominance distance to all points in any other cluster $C_k$ (where $C_k \neq C_i$), and select the minimum average distance among all such clusters:
	
	\begin{equation}
		b(i)_j^A = \min_{C_k \neq C_i} \left\{ \frac{1}{|C_k|} \sum_{v \in C_k} P(i, v)_j^A \right\}
	\end{equation}
	
	\begin{equation}
		b(i)_j^D = \min_{C_k \neq C_i} \left\{ \frac{1}{|C_k|} \sum_{v \in C_k} P(i, v)_j^D \right\}
	\end{equation}
	where $|C_k|$ denotes the number of samples in cluster $C_k$; $b(i)_j^A$ and $b(i)_j^D$ represent the inter-cluster distances corresponding to risk-averse and risk-seeking investors, respectively.
	
	Step 4: Calculate the SD-SC value for individual sample points and the overall clustering result:
	
	\begin{equation}
		s(i)_j^{A(D)} = \frac{b(i)_j^{A(D)} - a(i)_j^{A(D)}}{\max\{a(i)_j^{A(D)}, b(i)_j^{A(D)}\}}
	\end{equation}
	
	\begin{equation}
		SD\text{-}SC_j^{A(D)} = \frac{1}{n} \sum_{i=1}^n s(i)_j^{A(D)}
	\end{equation}
	where $n$ is the total number of stock samples; $SD\text{-}SC_j^A$ represents the overall SD-SC coefficient value for risk-averse investors, and $SD\text{-}SC_j^D$ represents the overall SD-SC coefficient value for risk-seeking investors.
	
	\subsection{SD-DBI Index}
	The Davies-Bouldin Index (DBI) is a classic internal clustering validity index used to measure the compactness and separation of clustering results. Traditional DBI calculations rely on Euclidean or Manhattan distances; however, these metrics cannot adequately reflect the performance of the stochastic dominance clustering model proposed in this paper. To address this limitation, this subsection proposes an improved clustering validity index based on the stochastic dominance distance, termed the SD-DBI Index, to better evaluate the clustering quality.
	
	The SD-DBI Index is derived from the traditional DBI by substituting the standard distance metric with the stochastic dominance distance. The specific calculation steps are as follows:
	
	Step 1: Define the distance metrics. Let $P(v, \mu_i)_j^A$ and $P(v, \mu_i)_j^D$ denote the $j$-th order stochastic dominance distances between a specific stock $v$ and its cluster center $\mu_i$. Consistent with Section 3.3, $\mu_i$ represents the mean return series of all stocks within cluster $i$.
	
	Step 2: Calculate the intra-cluster compactness $S_{i,j}$:
	
	\begin{equation}
		S_{i,j}^A = \frac{1}{|C_i|} \sum_{v \in C_i} P(v, \mu_i)_j^A
	\end{equation}
	
	\begin{equation}
		S_{i,j}^D = \frac{1}{|C_i|} \sum_{v \in C_i} P(v, \mu_i)_j^D
	\end{equation}
	where $\mu_i$ is the center of cluster $C_i$, and $|C_i|$ is the total number of samples in cluster $C_i$. $S_{i,j}^A$ and $S_{i,j}^D$ represent the intra-cluster compactness corresponding to risk-averse and risk-seeking investors under the $j$-th order SD, respectively.
	
	Step 3: Calculate the inter-cluster separation $M_{i,k,j}$:
	
	\begin{equation}
		M_{i,k,j}^A = P(\mu_i, \mu_k)_j^A
	\end{equation}
	
	\begin{equation}
		M_{i,k,j}^D = P(\mu_i, \mu_k)_j^D
	\end{equation}
	where $\mu_i$ and $\mu_k$ are the centers of cluster $C_i$ and cluster $C_k$, respectively. $M_{i,k,j}^A$ and $M_{i,k,j}^D$ represent the inter-cluster separation between the two cluster centers.
	
	Step 4: Calculate the inter-cluster similarity $R_{i,k,j}$ between cluster $C_i$ and cluster $C_k$:
	
	\begin{equation}
		R_{i,k,j}^{A(D)} = \frac{S_{i,j}^{A(D)} + S_{k,j}^{A(D)}}{M_{i,k,j}^{A(D)}}
	\end{equation}
	where $R_{i,k,j}^A$ and $R_{i,k,j}^D$ represent the pairwise inter-cluster similarity for risk-averse and risk-seeking investors, respectively.
	
	Step 5: Calculate the maximum similarity $R_{i,j}$ for cluster $C_i$:
	
	\begin{equation}
		R_{i,j}^{A(D)} = \max_{k \neq i} R_{i,k,j}^{A(D)}
	\end{equation}
	where $R_{i,j}^A$ and $R_{i,j}^D$ represent the maximum similarity values between cluster $C_i$ and its most similar neighboring cluster.
	
	Step 6: Calculate the overall SD-DBI Index:
	
	\begin{equation}
		SD\text{-}DBI_j^{A(D)} = \frac{1}{K} \sum_{i=1}^K R_{i,j}^{A(D)}
	\end{equation}
	where $K$ is the designated total number of clusters. $SD\text{-}DBI_j^A$ and $SD\text{-}DBI_j^D$ represent the overall SD-DBI coefficient values for risk-averse and risk-seeking investor scenarios, respectively. A smaller SD-DBI value indicates better clustering performance (i.e., high intra-cluster compactness and high inter-cluster separation).
	
	\section{Empirical Research}
	
	In this chapter, the empirical dataset comprises the weekly closing price data of the constituent stocks from two major market indices: the NASDAQ 100 Index (spanning from March 23, 2014, to March 24, 2024) and the CSI 100 Index (spanning from March 24, 2013, to March 24, 2024). Although the theoretical framework established in Chapter 3 encompasses up to the third-order stochastic dominance, this empirical section focuses exclusively on the first- and second-order SD clustering. In financial economics, the first two orders are generally sufficient to capture the fundamental investor preferences regarding non-satiation and risk aversion (or risk-seeking). Higher-order dominance relations often exhibit marginal empirical significance in broad market clustering and can overly complicate the interpretation of the results. Therefore, utilizing the first- and second-order stochastic dominance test statistics, we conduct stochastic dominance clustering analyses specifically tailored to risk-averse and risk-seeking investors.
	
	\subsection{Data Sources and Processing}
	\subsubsection{Data Sources}
	
	The NASDAQ 100 Index selects 100 of the largest domestic and international non-financial companies listed on the NASDAQ Stock Market based on market capitalization. Dominated primarily by high-tech and growth-oriented equities, it serves as a critical benchmark reflecting the overall trajectory of the US technology sector.
	
	Conversely, the CSI 100 Index comprises 100 large-capitalization and highly liquid stocks selected from the Shanghai and Shenzhen stock exchanges. It covers a diverse range of sectors, including finance, energy, consumer goods, and healthcare, effectively diversifying unsystematic risk while maintaining low liquidity risk. It is widely regarded as a representative barometer for the overall performance of blue-chip companies in China's A-share market.
	
	The historical trading data utilized in this study were retrieved from the Choice Financial Terminal (East Money Information Co., Ltd.), a comprehensive and professional financial database widely used in Chinese financial research.
	
	\subsubsection{Data Processing}
	
	To ensure the integrity of a balanced panel dataset, this study systematically excludes assets with incomplete trading records, specifically those unlisted at the onset of the sample period and those subject to prolonged suspensions. Following this rigorous filtering process, the finalized dataset curated for the subsequent clustering analysis retains 83 constituent stocks from the NASDAQ 100 and 76 from the CSI 100.
	
	This study utilizes logarithmic returns rather than simple discrete returns for modeling. Logarithmic returns possess several highly desirable statistical properties: notably, they satisfy time-additivity, which facilitates the calculation of multi-period returns, and they help mitigate the inherent positive skewness of asset prices. Therefore, logarithmic returns are highly suitable for long-term portfolio performance evaluation and robust stochastic modeling.
	
	Furthermore, to construct the return series, we aggregate daily closing prices into weekly intervals. Using weekly data effectively alleviates market microstructure noise (such as bid-ask bounce and asynchronous trading effects) prevalent in daily, high-frequency data. Simultaneously, it preserves sufficient observations to accurately capture the medium-to-long-term distributional characteristics and trends of the constituent stocks.
	
	\subsection{SD-K-means Clustering Analysis}
	
	Based on the theoretical framework established in Section 3.3, this section conducts K-means clustering analysis on the constituent stocks of the CSI 100 and the NASDAQ 100 indices. Depending on the investors' specific risk preferences and the applied order of stochastic dominance ($j$), the clustering algorithms are systematically categorized as follows:
	
	For risk-averse investors, the model adopts the ASD-K-means algorithm. Specifically, when utilizing the first-order ($j=1$), second-order ($j=2$), and third-order ($j=3$) stochastic dominance test statistics, the algorithms are denoted as FASD-K-means, SASD-K-means, and TASD-K-means, respectively.
	
	Conversely, for risk-seeking investors, the model applies the DSD-K-means algorithm. Accordingly, corresponding to $j=1, 2, \text{and } 3$, the algorithms are defined as FDSD-K-means, SDSD-K-means, and TDSD-K-means, respectively.
	
	While the full algorithmic suite incorporates up to the third order, the following empirical evaluations and results will primarily focus on the first- and second-order clustering scenarios, as they sufficiently capture the fundamental market behaviors of non-satiation and risk attitude.
	
	\subsubsection{ASD-K-means Clustering}
	
	This section first applies the first-order ASD-K-means clustering algorithm tailored for risk-averse investors. Building upon the first-order clustering results, we subsequently perform second-order ASD-K-means clustering to further refine the asset groupings. The empirical analysis utilizes the constituent stocks of both the NASDAQ 100 and the CSI 100 indices.
	
	Based on the SD-SC evaluation criteria discussed previously, the optimal number of clusters for the NASDAQ 100 stocks is determined to be 3, denoted as $K=3$. Figure 4.1 illustrates the heatmap of the stochastic dominance coefficient matrix generated from the first-order clustering. In this heatmap, the horizontal and vertical axes represent individual stocks, and each tiny square indicates the stochastic dominance distance between a given pair of stocks. A larger coefficient corresponds to a darker shade. The visual clustering pattern demonstrates excellent algorithmic performance. For instance, the first-order stochastic dominance coefficients among stocks within the third cluster are relatively low, creating a distinctively lighter square region along the diagonal. Conversely, the coefficients between these intra-cluster stocks and those outside the cluster are notably larger, as evidenced by the surrounding darker regions.
	
	\begin{figure}[htbp]
		\centering
		\includegraphics[width=0.8\textwidth]{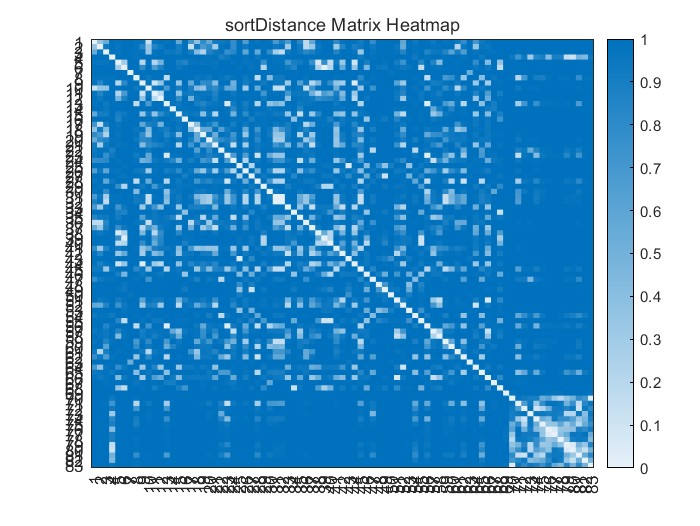}
		\caption{FASD-Kmeans Stochastic Dominance Coefficient Heatmap (NASDAQ)}
		\label{fig:1}
	\end{figure}
	
	Table 4.1 displays the constituency of each cluster following the first-order ASD-K-means clustering of the NASDAQ stocks. Subsequently, the calculated center of each cluster is selected as its representative to compare the first-order stochastic dominance relationships among the three clusters.

	\begin{table}[htbp]
		\centering
		\caption{FASD-K-means Clustering Results (NASDAQ)}
		\setlength{\tabcolsep}{4pt} 
		\begin{tabular}{c p{10cm} c}
			\hline
			Cluster & Stock Names & Count \\
			\hline
			1 & Apple,\ Adobe,\ Analog Devices,\ Automatic Data Processing (ADP),\ Autodesk,\ Applied Materials,\ AMD,\ Amgen,\ Amazon,\ ANSYS,\ ASML,\ Broadcom,\ AstraZeneca (US ADR),\ Biogen,\ Booking Holdings,\ Baker Hughes,\ Cadence Design Systems,\ CDW Corp,\ Charter Communications,\ Copart,\ CoStar Group,\ Cintas,\ Cognizant,\ Dollar Tree,\ DexCom,\ Electronic Arts,\ Diamondback Energy,\ Fastenal,\ Fortinet,\ Gilead Sciences,\ Alphabet C (GOOG),\ Alphabet A (GOOGL),\ IDEXX Laboratories,\ Illumina,\ Intel,\ Intuit,\ Intuitive Surgical,\ KLA Corp,\ Lam Research,\ Lululemon Athletica,\ Marriott International,\ Microchip Technology,\ MercadoLibre,\ Meta Platforms,\ Monster Beverage,\ Marvell Technology,\ Microsoft,\ Micron Technology,\ Netflix,\ NVIDIA,\ NXP Semiconductors,\ Old Dominion Freight Line,\ ON Semiconductor,\ O'Reilly Automotive,\ Palo Alto Networks,\ PACCAR,\ Qualcomm,\ Regeneron Pharmaceuticals,\ Ross Stores,\ Sirius XM,\ Synopsys,\ T-Mobile US,\ Tesla,\ Take-Two Interactive,\ Texas Instruments,\ Vertex Pharmaceuticals,\ Walgreens Boots Alliance,\ Workday & 68 \\
			2 & PepsiCo & 1 \\
			3 & American Electric Power,\ Coca-Cola Europacific Partners,\ Comcast,\ Costco Wholesale,\ Cisco Systems,\ Exelon,\ Honeywell International,\ Keurig Dr Pepper,\ Mondelez International,\ Paychex,\ Roper Technologies,\ Starbucks,\ Verisk Analytics,\ Xcel Energy & 14 \\
			\hline
		\end{tabular}
		\label{tab:4-1}
	\end{table}
	
	\begin{table}[htbp]
		\centering
		\caption{FASD-Kmeans Stochastic Dominance Relationships Among Cluster Centers (NASDAQ)}
		\begin{tabular}{l p{3cm} p{3cm} p{3cm}}
			\hline
			& Cluster 1 ($C_1$) & Cluster 2 ($C_2$) & Cluster 3 ($C_3$) \\
			\hline
			\textbf{Cluster 1} & - & $\succ_1$ & $\succ_1$ \\
			\textbf{Cluster 2} & - & - & $\succ_1$ \\
			\textbf{Cluster 3} & $\succ_1$ & $\succ_1$ & - \\
			\hline
			\multicolumn{4}{p{12cm}}{\small \textit{Note:} $\succ_1$ indicates dominance, $\prec_1$ indicates being dominated, and $\succ_1 \prec_1$ indicates no clear dominance relationship.} \\
		\end{tabular}
		\label{tab:4-2}
	\end{table}
	
	Observations from Table 4.2 indicate that the first cluster center strictly dominates the second in the first order, denoted as $C_1 \succ_1 C_2$. However, no definitive first-order stochastic dominance relationship can be established between $C_1$ and $C_3$, or between $C_2$ and $C_3$. To resolve this ambiguity, a granular comparison is conducted between the individual constituent stocks of the third cluster and the centers $C_1$ and $C_2$. A comprehensive analysis reveals that five specific stocks within the third cluster—Costco Wholesale, Keurig Dr Pepper, Mondelez International, Paychex, and Xcel Energy—first-order stochastically dominate $C_2$. Among these, Costco Wholesale, Paychex, and Xcel Energy exhibit no clear first-order dominance relationship with $C_1$, whereas all other remaining stocks in the third cluster are strictly dominated by $C_1$ at the first-order level. Consequently, this study extracts these three specific non-dominated stocks—Costco Wholesale, Paychex, and Xcel Energy—and merges them with the constituents of the first cluster to perform a subsequent second-order stochastic dominance clustering. This targeted refinement aims to further isolate the asset pools and determine the optimal stock groupings tailored to distinct investor types.
	
	It is found that there is no significant first-order stochastic dominance relationship among the 76 stocks of CSI 100 index, indicating that there is no superiority or inferiority of these stocks under the first-order stochastic dominance condition, which needs to be further analysed by higher-order stochastic dominance Kmeans clustering.
	
	For risk-averse investors, the SASD-K-means clustering algorithm is applied to the refined pools comprising 71 NASDAQ constituent stocks and 76 CSI 100 constituent stocks. By maximizing the SD-SC evaluation metric, the optimal number of clusters is determined to be three for the NASDAQ subset and two for the CSI 100 subset. The corresponding second-order stochastic dominance coefficient heatmaps are presented in Figure 4.2. Bypassing basic coordinate definitions, the visual evidence in the left panel clearly partitions the NASDAQ stocks into three distinct classes based on the magnitude of their stochastic dominance coefficients. Concurrently, the right panel depicts the CSI 100 constituents, revealing that the stocks within the second cluster, located in the bottom-right corner, exhibit remarkably large stochastic dominance coefficients when compared against other stocks. This strong visual contrast indicates a highly pronounced second-order stochastic dominance relationship.
			
	\begin{figure}[htbp]
		\subfloat[]{\includegraphics[width=0.45\textwidth]{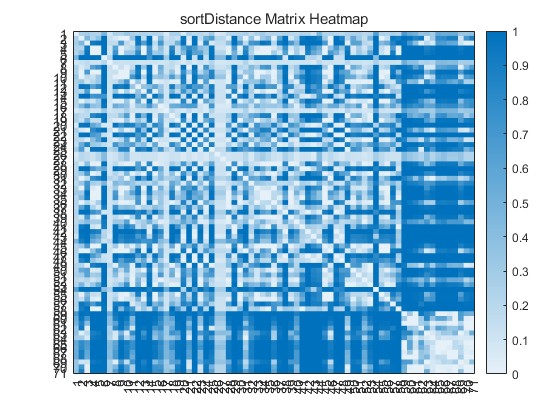}}
		\hfill 
		\subfloat[]{\includegraphics[width=0.45\textwidth]{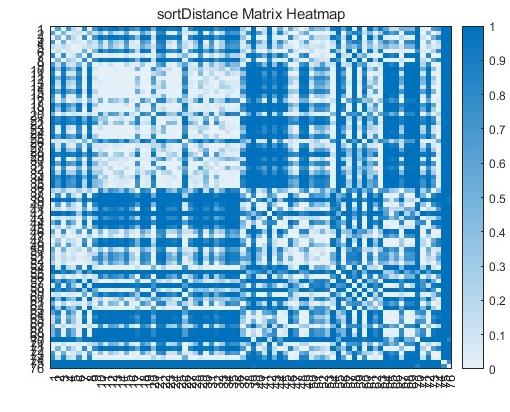}}
		\caption{SASD-Kmeans Stochastic Dominance Coefficient Heatmap}
		\label{fig:2}
	\end{figure}
	
	A comparison of the stochastic dominance relationships among the centers of the three NASDAQ clusters reveals no definitive second-order stochastic dominance between the second and third clusters. However, both of these clusters dominate the first cluster in the second order, expressed mathematically as $C_2 \equiv C_3$, $C_2 \succ_2 C_1$, and $C_3 \succ_2 C_1$. Consequently, this study merges the second and third clusters into a single unified class, with the consolidated clustering results presented in Table 4.3. Furthermore, Table 4.4 details the specific constituency of each stock class following the CSI 100 clustering.
	
	By calculating the cross-sectional mean of all samples within the second cluster of Table 4.3 to serve as the updated cluster center, a subsequent comparison reveals that the second cluster strictly dominates the first in the second order, denoted as $C_2 \succ_2 C_1$. Consequently, this study ultimately selects the 13 constituent stocks from the second NASDAQ cluster for investment and final portfolio construction. Following a parallel evaluation of the two cluster centers detailed in Table 4.4, the results identically demonstrate that the second cluster strictly dominates the first, expressed mathematically as $C_2 \succ_2 C_1$. Therefore, for risk-averse investors, allocating capital to the two specific constituent stocks of this optimal CSI 100 cluster—China Yangtze Power and the Industrial and Commercial Bank of China—yields a substantially higher expected utility.
	
	\begin{table}[htbp]
		\centering
		\caption{SASD-K-means Clustering Results (NASDAQ)}
		\setlength{\tabcolsep}{4pt} 
		\begin{tabular}{c p{10cm} c}
			\hline
			Cluster & Stock Names & Count \\
			\hline
			1 & Apple, \ Adobe, \ Analog Devices, \ Autodesk, \ Applied Materials, \ AMD, \ Amazon, \ ANSYS, \ ASML, \ Broadcom, \ AstraZeneca (US ADR), \ Biogen, \ Booking Holdings, \ Baker Hughes, \ Charter Communications, \ Copart, \ CoStar Group, \ Cognizant, \ Dollar Tree, \ DexCom, \ Electronic Arts, \ Diamondback Energy, \ Fastenal, \ Fortinet, \ Gilead Sciences, \ Alphabet C (GOOG), \ Alphabet A (GOOGL), \ IDEXX Laboratories, \ Illumina, \ Intel, \ Intuit, \ Intuitive Surgical, \ KLA Corp, \ Lam Research, \ Lululemon Athletica, \ Marriott International, \ Microchip Technology, \ MercadoLibre, \ Meta Platforms, \ Monster Beverage, \ Marvell Technology, \ Micron Technology, \ Netflix, \ NVIDIA, \ NXP Semiconductors, \ Old Dominion Freight Line, \ ON Semiconductor, \ Palo Alto Networks, \ PACCAR, \ Qualcomm, \ Regeneron Pharmaceuticals, \ Ross Stores, \ Sirius XM, \ Tesla, \ Take-Two Interactive, \ Vertex Pharmaceuticals, \ Walgreens Boots Alliance, \ Workday & 58 \\
			2 & Costco Wholesale, \ Xcel Energy, \ Automatic Data Processing (ADP), \ Amgen, \ Cadence Design Systems, \ CDW Corp, \ Cintas, \ Microsoft, \ O'Reilly Automotive, \ Paychex, \ Synopsys, \ T-Mobile US, \ Texas Instruments & 13 \\
			\hline
		\end{tabular}
		\label{tab:4-3}
	\end{table}
	
	\begin{table}[htbp]
		\centering
		\caption{SASD-K-means Clustering Results (CSI 100)}
		\setlength{\tabcolsep}{4pt} 
		\begin{tabular}{c p{10cm} c}
			\hline
			Cluster & Stock Names & Count \\
			\hline
			1 & China Vanke, \ ZTE Corporation, \ TCL Technology, \ Eastern Shenghong, \ Weichai Power, \ Gree Electric Appliances, \ Changchun High-Tech, \ BOE Technology, \ Qinghai Salt Lake Industry, \ Unisplendour, \ Focus Media, \ Unigroup Guoxin Microelectronics, \ TCL Zhonghuan Renewable Energy, \ iFLYTEK, \ Goertek, \ Oriental Yuhong, \ SF Holding, \ NAURA Technology, \ Glodon, \ Hikvision, \ Ganfeng Lithium, \ Tianqi Lithium, \ Luxshare Precision, \ Rongsheng Petrochemical, \ 37 Interactive Entertainment, \ BYD, \ LB Group, \ Satellite Chemical, \ EVE Energy, \ Aier Eye Hospital, \ Zhifei Biological, \ Inovance Technology, \ Walvax Biotechnology, \ Sungrow Power Supply, \ Zhongji Innolight, \ Tigermed, \ Baotou Steel, \ Baoshan Iron \& Steel (Baosteel), \ Sinopec, \ CITIC Securities, \ SANY Heavy Industry, \ China Merchants Bank, \ Poly Developments, \ China Unicom, \ TBEA, \ China Northern Rare Earth, \ China Jushi, \ Hengrui Medicine, \ Wanhua Chemical, \ Hengli Petrochemical, \ NARI Technology, \ Hualu Hengsheng, \ Pien Tze Huang, \ Tongwei, \ Kweichow Moutai, \ Shandong Gold, \ Hundsun Technologies, \ Conch Cement, \ Yonyou Network, \ Haier Smart Home, \ Wingtech Technology, \ AECC Aviation Power, \ LONGi Green Energy, \ China Shenhua Energy, \ Ping An Insurance, \ China Railway Group, \ Aluminum Corporation of China (Chalco), \ China State Construction Engineering (CSCEC), \ CRRC, \ PetroChina, \ China Tourism Group Duty Free, \ Zijin Mining, \ COSCO SHIPPING Holdings, \ CMOC Group & 74 \\
			2 & China Yangtze Power, \ Industrial and Commercial Bank of China (ICBC) & 2 \\
			\hline
		\end{tabular}
		\label{tab:4-4}
	\end{table}
	
	\subsubsection{DSD-K-means Clustering}
	
	Subsequently, stock K-means clustering is performed for risk-seeking investors, first performing first-order K-means clustering (FDSD-K-means clustering), and on the basis of the first-order clustering, continuing to perform second-order K-means clustering (SDSD-K-means clustering). Both theoretical and experimental results indicate that the results of FASD-K-means clustering and FDSD-K-means clustering are the same. Therefore, for risk-seeking investors, this paper still performs SDSD-K-means clustering on the 71 NASDAQ constituent stocks and 76 CSI 100 constituent stocks.
	
	\begin{figure}[htbp]
		\subfloat[]{\includegraphics[width=0.45\textwidth]{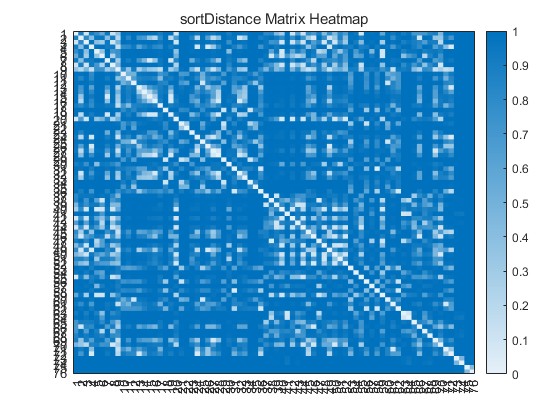}}
		\hfill 
		\subfloat[]{\includegraphics[width=0.45\textwidth]{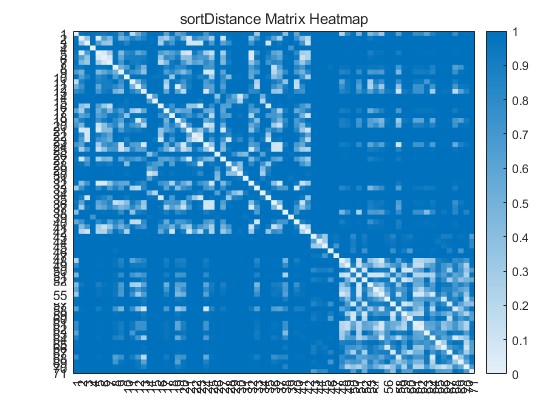}}
		\caption{SDSD-Kmeans Stochastic Dominance Coefficient Heatmap}
		\label{fig:3}
	\end{figure}
	
	The optimal number of clusters for NASDAQ is 4, and the optimal number of clusters for CSI 100 stocks is 7. The left chart in Figure 4-3 is the stochastic dominance coefficient heatmap for NASDAQ after clustering, and the right chart is for the CSI 100 constituents. In the figure, the horizontal and vertical axes represent each individual stock, and each small square represents the stochastic dominance coefficient value between two stocks; the larger the value, the darker the color of the square. It can be seen from the figure that the effect after clustering the two datasets is very good, and the stocks are clustered into several classes with obvious differences in stochastic dominance coefficients.
	
	The specific stock clustering situation is shown in Table 4-5 and Table 4-6. Comparing the second-order stochastic dominance relationships among the centers of each class of stocks in Table 4-5, it is found that $C_1 \succ_2 C_2$, $C_1 \succ_2 C_3$, $C_1 \succ_2 C_4$. For risk-seeking investors, the 42 stocks in the first class can be selected from the NASDAQ constituents for subsequent investment portfolios. Comparing the second-order stochastic dominance relationships among the cluster centers in Table 4-6, the results show that the center of Class 1 stochastically dominates the centers of all other classes in the second order. Therefore, the 72 stocks in Class 1 of Table 4-6 are selected for the investment portfolio.
	
	\begin{table}[htbp]
		\centering
		\caption{SDSD-K-means Clustering Results (NASDAQ)}
		\setlength{\tabcolsep}{4pt} 
		\begin{tabular}{c p{10cm} c}
			\hline
			Cluster & Stock Names & Count \\
			\hline
			1 & Adobe, \ Autodesk, \ Applied Materials, \ AMD, \ Amazon, \ ASML, \ Broadcom, \ Biogen, \ Booking Holdings, \ Baker Hughes, \ Cadence Design Systems, \ CoStar Group, \ Dollar Tree, \ DexCom, \ Diamondback Energy, \ Fortinet, \ IDEXX Laboratories, \ Illumina, \ Intel, \ Intuitive Surgical, \ KLA Corp, \ Lam Research, \ Lululemon Athletica, \ Marriott International, \ Microchip Technology, \ MercadoLibre, \ Meta Platforms, \ Marvell Technology, \ Micron Technology, \ Netflix, \ NVIDIA, \ NXP Semiconductors, \ Old Dominion Freight Line, \ ON Semiconductor, \ Palo Alto Networks, \ Qualcomm, \ Regeneron Pharmaceuticals, \ Synopsys, \ Tesla, \ Take-Two Interactive, \ Vertex Pharmaceuticals, \ Workday & 42 \\
			2 & Automatic Data Processing (ADP), \ Costco Wholesale, \ Paychex, \ Walgreens Boots Alliance & 4 \\
			3 & Xcel Energy & 1 \\
			4 & Apple, \ Analog Devices, \ Amgen, \ ANSYS, \ AstraZeneca (US ADR), \ CDW Corp, \ Charter Communications, \ Copart, \ Cintas, \ Cognizant, \ Electronic Arts, \ Fastenal, \ Gilead Sciences, \ Alphabet C (GOOG), \ Alphabet A (GOOGL), \ Intuit, \ Monster Beverage, \ Microsoft, \ O'Reilly Automotive, \ PACCAR, \ Ross Stores, \ Sirius XM, \ T-Mobile US, \ Texas Instruments & 24 \\
			\hline
		\end{tabular}
		\label{tab:4-5}
	\end{table}

	\begin{table}[htbp]
		\centering
		\caption{SDSD-K-means Clustering Results (CSI 100)}
		\setlength{\tabcolsep}{4pt} 
		\begin{tabular}{c p{10cm} c}
			\hline
			Cluster & Stock Names & Count \\
			\hline
			1 & ZTE Corporation, \ Eastern Shenghong, \ Changchun High-Tech, \ Unisplendour, \ Focus Media, \ Unigroup Guoxin Microelectronics, \ TCL Zhonghuan Renewable Energy, \ iFLYTEK, \ Goertek, \ Oriental Yuhong, \ SF Holding, \ NAURA Technology, \ Glodon, \ Ganfeng Lithium, \ Tianqi Lithium, \ Luxshare Precision, \ Rongsheng Petrochemical, \ 37 Interactive Entertainment, \ BYD, \ LB Group, \ Satellite Chemical, \ EVE Energy, \ Aier Eye Hospital, \ Zhifei Biological, \ Inovance Technology, \ Walvax Biotechnology, \ Sungrow Power Supply, \ Zhongji Innolight, \ Tigermed, \ China Northern Rare Earth, \ China Jushi, \ Hengli Petrochemical, \ Hualu Hengsheng, \ Tongwei, \ Hundsun Technologies, \ Yonyou Network, \ Wingtech Technology, \ AECC Aviation Power, \ LONGi Green Energy, \ China Tourism Group Duty Free, \ COSCO SHIPPING Holdings, \ CMOC Group, \ China Vanke, \ TCL Technology, \ BOE Technology, \ Qinghai Salt Lake Industry, \ Hikvision, \ Baotou Steel, \ CITIC Securities, \ Poly Developments, \ China Unicom, \ TBEA, \ Wanhua Chemical, \ NARI Technology, \ Pien Tze Huang, \ Shandong Gold, \ China Railway Group, \ Aluminum Corporation of China (Chalco), \ China State Construction Engineering (CSCEC), \ CRRC, \ Zijin Mining, \ Weichai Power, \ Gree Electric Appliances, \ Baoshan Iron \& Steel (Baosteel), \ SANY Heavy Industry, \ China Merchants Bank, \ Hengrui Medicine, \ Kweichow Moutai, \ Conch Cement, \ Haier Smart Home, \ China Shenhua Energy, \ Ping An Insurance & 72 \\
			2 & Sinopec, \ PetroChina & 2 \\
			3 & China Yangtze Power & 1 \\
			4 & Industrial and Commercial Bank of China (ICBC) & 1 \\
			\hline
		\end{tabular}
		\label{tab:4-6}
	\end{table}
	
	\subsubsection{Model Performance Evaluation}
	
	The values of the classic K-means clustering results are compared with those of the SD clustering models. To ensure comparability, the number of clusters selected for the classic clustering models is kept consistent with that of the SD models.
	
	\begin{table}[htbp]
		\centering
		\caption{Comparison of SD-SC Values Between Classical and SD Clustering Models}
		\setlength{\tabcolsep}{12pt} 
		\begin{tabular}{ccccc}
			\toprule
			\multirow{2}{*}{} & \multicolumn{2}{c}{NASDAQ Dataset} & \multicolumn{2}{c}{CSI 100 Dataset} \\
			\cmidrule(lr){2-3} \cmidrule(lr){4-5}
			& SD & Classical & SD & Classical \\
			\midrule
			FASD(DSD)-K-means & 0.1495 & 0.0546 & & \\
			SASD-K-means      & 0.3838 & -0.3760 & 0.4827 & 0.0894 \\
			SDSD-K-means      & 0.1424 & -0.1541 & 0.1872 & -0.1645 \\
			\bottomrule
		\end{tabular}
		\label{tab:4-7}
	\end{table}
	
	\vspace{2em} 
	
	\begin{table}[htbp]
		\centering
		\caption{Comparison of SD-DBI Values Between Classical and SD Clustering Models}
		\setlength{\tabcolsep}{12pt}
		\begin{tabular}{ccccc}
			\toprule
			\multirow{2}{*}{} & \multicolumn{2}{c}{NASDAQ Dataset} & \multicolumn{2}{c}{CSI 100 Dataset} \\
			\cmidrule(lr){2-3} \cmidrule(lr){4-5}
			& SD & Classical & SD & Classical \\
			\midrule
			FASD(DSD)-K-means & 0.6674 & 1.5815 & & \\
			SASD-K-means      & 1.6135 & 2.6940 & 1.8048 & 0.9175 \\
			SDSD-K-means      & 1.0787 & 2.2599 & 2.2029 & 1.3338 \\
			\bottomrule
		\end{tabular}
		\label{tab:4-8}
	\end{table}
	
	Table 4-7 presents the SD-SC values, and Table 4-8 presents the SD-DBI values. As seen in Table 4-7, the SD-SC values for all ASD-K-means and DSD-K-means clustering models across both datasets are higher than those of the classic clustering results. Table 4-8 shows that the SD-DBI values for the first-order and second-order ASD-K-means and DSD-K-means models on the NASDAQ dataset are all lower than the classic clustering results, verifying the quality of the clustering. However, the SD-DBI values for the CSI 100 dataset are all higher than the classic clustering results, which may be due to the weaker applicability of the SD-DBI index on this specific dataset.
	
	A comprehensive analysis combining both indices reveals that higher SD-SC coefficient values reflect better intra-class compactness and inter-class separation, while lower DBI values verify the rationality and effectiveness of the clustering results. Therefore, the index results demonstrate that the stochastic dominance K-means clustering models proposed in this paper exhibit superior performance in stock clustering tasks targeting different investment preferences. Consequently, it can be concluded that the ASD-K-means and DSD-K-means clustering models possess significant advantages in applications regarding stock clustering based on investor preferences, providing more reliable references for investment decision-making.
	
	\subsection{SD-Hierarchical Clustering Analysis}
	
	Based on the theoretical framework established in Section 3.4, this section conducts hierarchical clustering analysis on the constituent stocks of the CSI 100 and the NASDAQ 100 indices. Depending on the investors' specific risk preferences and the applied order of stochastic dominance, the clustering algorithms are systematically categorized as follows:
	
	For risk-averse investors, the model adopts the ASD-Hierarchical algorithm. Specifically, when utilizing the first-order, second-order, and third-order stochastic dominance test statistics, the algorithms are denoted as FASD-Hierarchical, SASD-Hierarchical, and TASD-Hierarchical, respectively.
	
	Conversely, for risk-seeking investors, the model applies the DSD-Hierarchical algorithm. Accordingly, corresponding to the application of the first-order, second-order, and third-order statistics, the algorithms are defined as FDSD-Hierarchical, SDSD-Hierarchical, and TDSD-Hierarchical, respectively.
	
	While the full algorithmic suite incorporates up to the third order, the following empirical evaluations and results will primarily focus on the first- and second-order clustering scenarios, as they sufficiently capture the fundamental market behaviors of non-satiation and risk attitude.
	
	\subsubsection{ASD-Hierarchical Clustering}
	
	This paper first conducts first-order stock hierarchical clustering (FASD-Hierarchical clustering) for risk-averse investors, followed by second-order hierarchical clustering (SASD-Hierarchical clustering) based on the first-order results. The FASD-Hierarchical clustering is performed on NASDAQ index constituents and CSI 100 constituents respectively. The optimal number of clusters selected for NASDAQ stocks is 20. Figure 4-4 plots the heatmap of stochastic dominance coefficients after clustering. In the figure, the horizontal and vertical axes represent individual stocks, and each small square represents the stochastic dominance coefficient value between two stocks; larger values correspond to darker colors. The figure reveals a clear block structure in the stochastic dominance coefficients between stocks after clustering. Compared to the FASD-K-means and FDSD-K-means results, the stochastic dominance relationships within and between classes are more distinct after FASD-Hierarchical and FDSD-Hierarchical clustering.
	
	\begin{figure}[htbp]
		\centering
		\includegraphics[width=0.8\textwidth]{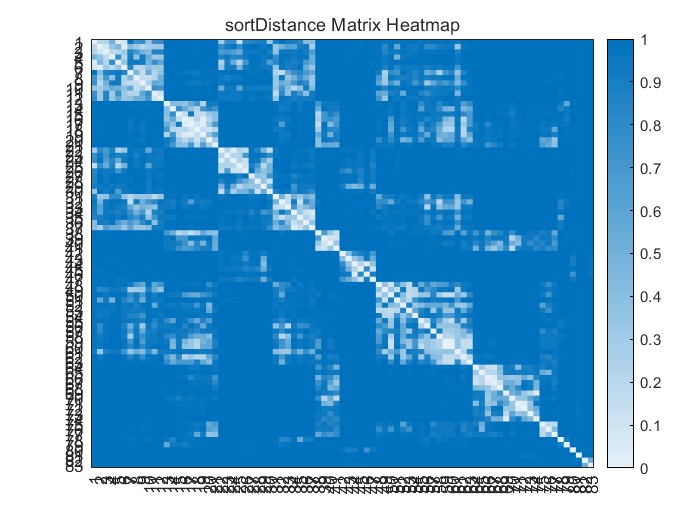}
		\caption{FASD-Hierarchical Stochastic Dominance Coefficient Heatmap (NASDAQ)}
		\label{fig:4}
	\end{figure}
	
	Subsequently, the center of each class of stocks is selected as a representative to compare the first-order stochastic dominance relationships between cluster centers. Based on the first-order stochastic dominance results of the cluster centers, this paper ultimately selects 75 stocks (excluding those in classes 3, 6, 16, and 17) for second-order hierarchical clustering to obtain more distinct stochastic dominance relationships and select the optimal stocks for an investment portfolio. The details of each class of NASDAQ stocks after first-order hierarchical clustering are shown in Table 4-9.
	
	\begin{table}[htbp]
		\centering
		\caption{FASD-Hierarchical Clustering Results (NASDAQ)}
		\setlength{\tabcolsep}{4pt} 
		\begin{tabular}{c p{10cm} c}
			\hline
			Cluster & Stock Names & Count \\
			\hline
			1 & Autodesk,\ Amazon,\ ASML,\ Broadcom,\ KLA Corp,\ Meta Platforms & 6 \\
			2 & CoStar Group,\ IDEXX Laboratories,\ Intuitive Surgical,\ Microchip Technology,\ Old Dominion Freight Line,\ Take-Two Interactive & 6 \\
			3 & Gilead Sciences,\ Sirius XM & 2 \\
			4 & Amgen,\ AstraZeneca (US ADR),\ Cognizant,\ Fastenal,\ PACCAR,\ T-Mobile US,\ Texas Instruments & 7 \\
			5 & Applied Materials,\ Fortinet,\ Lam Research,\ Lululemon Athletica,\ Palo Alto Networks & 5 \\
			6 & Baker Hughes,\ Illumina,\ Marvell Technology,\ Workday & 4 \\
			7 & Booking Holdings,\ Dollar Tree,\ Intel,\ NXP Semiconductors,\ Qualcomm,\ Regeneron Pharmaceuticals,\ Vertex Pharmaceuticals & 7 \\
			8 & Coca-Cola Europacific Partners,\ Comcast,\ Cisco Systems,\ Starbucks & 4 \\
			9 & Micron Technology,\ ON Semiconductor & 2 \\
			10 & DexCom,\ Diamondback Energy,\ MercadoLibre,\ Netflix & 4 \\
			11 & Apple,\ Adobe,\ ANSYS,\ Cadence Design Systems,\ CDW Corp,\ Intuit,\ Synopsys & 7 \\
			12 & Analog Devices,\ Charter Communications,\ Copart,\ Electronic Arts,\ Alphabet C (GOOG),\ Alphabet A (GOOGL),\ Marriott International,\ Monster Beverage,\ Ross Stores & 9 \\
			13 & Automatic Data Processing (ADP),\ Costco Wholesale,\ Paychex,\ Roper Technologies,\ Verisk Analytics & 5 \\
			14 & American Electric Power,\ Exelon,\ Honeywell International,\ Keurig Dr Pepper,\ Mondelez International,\ Xcel Energy & 6 \\
			15 & Cintas,\ Microsoft,\ O'Reilly Automotive & 3 \\
			16 & Biogen & 1 \\
			17 & Walgreens Boots Alliance & 1 \\
			18 & NVIDIA & 1 \\
			19 & PepsiCo & 1 \\
			20 & AMD,\ Tesla & 2 \\
			\hline
		\end{tabular}
		\label{tab:4-9}
	\end{table}
	
	Research indicates that there are no obvious first-order stochastic dominance relationships among the CSI 100 stocks. Therefore, second-order hierarchical clustering (SASD-Hierarchical clustering) is performed directly on the 76 CSI 100 stocks.
	
	\begin{figure}[htbp]
		\subfloat[]{\includegraphics[width=0.45\textwidth]{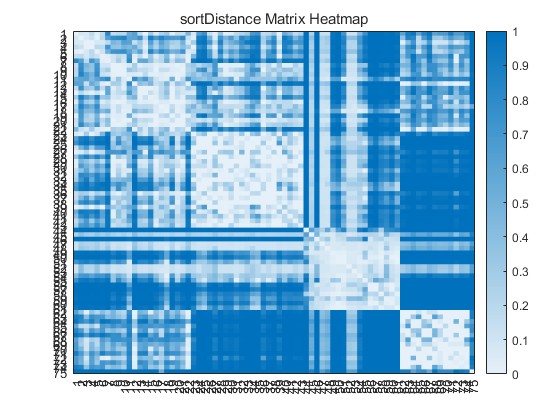}}
		\hfill 
		\subfloat[]{\includegraphics[width=0.45\textwidth]{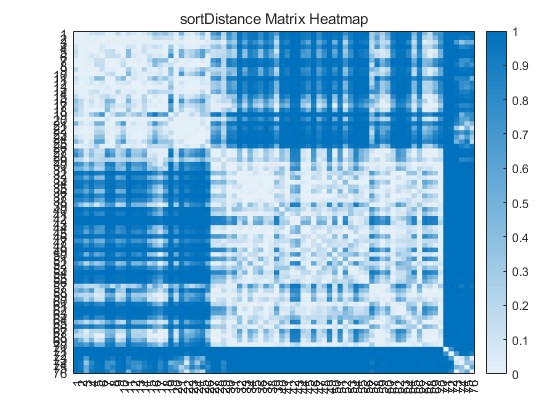}}
		\caption{SASD-Hierarchical Stochastic Dominance Coefficient Heatmap}
		\label{fig:5}
	\end{figure}

	For risk-averse investors, SASD-Hierarchical clustering is conducted on the 75 NASDAQ constituent stocks and the 76 CSI 100 constituent stocks. The optimal number of clusters for NASDAQ is 7, while for CSI 100 stocks, the silhouette coefficient is maximized at approximately 0.72 when clustered into 5 classes. The left chart in Figure 4-5 is the heatmap of stochastic dominance coefficients for NASDAQ after clustering. It can be seen that stocks are clustered into 7 distinct classes based on the magnitude of stochastic dominance coefficients; the right chart shows CSI 100 constituents, indicating relatively poor differentiation in second-order stochastic dominance coefficients.
	
	\begin{table}[htbp]
		\centering
		\caption{SASD-Hierarchical Clustering Results (NASDAQ)}
		\setlength{\tabcolsep}{4pt} 
		\begin{tabular}{c p{10cm} c}
			\hline
			Cluster & Stock Names & Count \\
			\hline
			1 & ANSYS,\ Cadence Design Systems,\ CDW Corp,\ Intuit,\ Synopsys & 5 \\
			2 & Analog Devices,\ Amgen,\ AstraZeneca (US ADR),\ Coca-Cola Europacific Partners,\ Comcast,\ Copart,\ Cisco Systems,\ Cognizant,\ Electronic Arts,\ Exelon,\ Fastenal,\ Keurig Dr Pepper,\ Monster Beverage,\ PACCAR,\ Ross Stores,\ Starbucks,\ T-Mobile US & 17 \\
			3 & Adobe,\ Autodesk,\ Applied Materials,\ ASML,\ Broadcom,\ Booking Holdings,\ Charter Communications,\ Dollar Tree,\ IDEXX Laboratories,\ Intel,\ KLA Corp,\ Lam Research,\ Lululemon Athletica,\ Marriott International,\ Meta Platforms,\ NXP Semiconductors,\ Old Dominion Freight Line,\ Qualcomm,\ Regeneron Pharmaceuticals,\ Take-Two Interactive,\ Vertex Pharmaceuticals & 21 \\
			4 & Tesla & 1 \\
			5 & Apple,\ AMD,\ Amazon,\ CoStar Group,\ DexCom,\ Diamondback Energy,\ Fortinet,\ Alphabet C (GOOG),\ Alphabet A (GOOGL),\ Intuitive Surgical,\ Microchip Technology,\ MercadoLibre,\ Micron Technology,\ Netflix,\ NVIDIA,\ ON Semiconductor,\ Palo Alto Networks & 17 \\
			6 & Automatic Data Processing (ADP),\ American Electric Power,\ Costco Wholesale,\ Cintas,\ Honeywell International,\ Mondelez International,\ Microsoft,\ O'Reilly Automotive,\ Paychex,\ Roper Technologies,\ Texas Instruments,\ Verisk Analytics,\ Xcel Energy & 13 \\
			7 & PepsiCo & 1 \\
			\hline
		\end{tabular}
		\label{tab:4-10}
	\end{table}
	
	Comparing the stochastic dominance relationships of the centers of each class in Table 4-10, it is found that $C_2 \equiv C_6 \equiv C_7$, $C_2(C_6, C_7) \succ_2 C_4$, and $C_2(C_6, C_7) \succ_2 C_5$; $C_1 \equiv C_3$, and $C_1(C_3) \succ_2 C_4$. Therefore, this paper merges classes 2, 6, and 7 to form a new class; classes 1 and 3 are merged to form a new second class. The second-order stochastic dominance relationship between the two new classes is then compared. Specifics are shown in Table 4-11.
	
	\begin{table}[htbp]
		\centering
		\caption{Merged SASD-Hierarchical Clustering Results (NASDAQ)}
		\setlength{\tabcolsep}{4pt} 
		\begin{tabular}{c p{10cm} c}
			\hline
			Cluster & Stock Names & Count \\
			\hline
			1 & Analog Devices,\ Amgen,\ AstraZeneca (US ADR),\ Coca-Cola Europacific Partners,\ Comcast,\ Copart,\ Cisco Systems,\ Cognizant,\ Electronic Arts,\ Exelon,\ Fastenal,\ Keurig Dr Pepper,\ Monster Beverage,\ PACCAR,\ Ross Stores,\ Starbucks,\ T-Mobile US,\ Texas Instruments,\ Verisk Analytics,\ Xcel Energy,\ Automatic Data Processing (ADP),\ American Electric Power,\ Costco Wholesale,\ Cintas,\ Honeywell International,\ Mondelez International,\ Microsoft,\ O'Reilly Automotive,\ Paychex,\ Roper Technologies,\ PepsiCo & 31 \\
			2 & ANSYS,\ Cadence Design Systems,\ CDW Corp,\ Intuit,\ Synopsys,\ Adobe,\ Autodesk,\ Applied Materials,\ ASML,\ Broadcom,\ Booking Holdings,\ Charter Communications,\ Dollar Tree,\ IDEXX Laboratories,\ Intel,\ KLA Corp,\ Lam Research,\ Lululemon Athletica,\ Marriott International,\ Meta Platforms,\ NXP Semiconductors,\ Old Dominion Freight Line,\ Qualcomm,\ Regeneron Pharmaceuticals,\ Take-Two Interactive,\ Vertex Pharmaceuticals & 26 \\
			\hline
		\end{tabular}
		\label{tab:4-11}
	\end{table}
	
	Taking the mean value of the samples in each class in Table 4-11 as the cluster center, and comparing the second-order stochastic dominance relationship between the centers of the two classes, it is found that $C_1 \succ_2 C_2$. For the NASDAQ dataset, risk-averse investors can select the 31 stocks in the first class for investment.
	
	\begin{table}[htbp]
		\centering
		\caption{SASD-Hierarchical Clustering Results (CSI 100)}
		\setlength{\tabcolsep}{4pt} 
		\begin{tabular}{c p{10cm} c}
			\hline
			Cluster & Stock Names & Count \\
			\hline
			1 & China Vanke, \ TCL Technology, \ Gree Electric Appliances, \ BOE Technology, \ Hikvision, \ BYD, \ Baoshan Iron \& Steel (Baosteel), \ CITIC Securities, \ SANY Heavy Industry, \ Poly Developments, \ China Unicom, \ TBEA, \ Hengrui Medicine, \ Wanhua Chemical, \ NARI Technology, \ Hualu Hengsheng, \ Pien Tze Huang, \ Shandong Gold, \ Conch Cement, \ Haier Smart Home, \ China Shenhua Energy, \ Ping An Insurance, \ China Railway Group, \ China State Construction Engineering (CSCEC), \ CRRC, \ Zijin Mining & 26 \\
			2 & ZTE Corporation, \ Eastern Shenghong, \ Weichai Power, \ Changchun High-Tech, \ Qinghai Salt Lake Industry, \ Unisplendour, \ Focus Media, \ Unigroup Guoxin Microelectronics, \ TCL Zhonghuan Renewable Energy, \ iFLYTEK, \ Goertek, \ Oriental Yuhong, \ SF Holding, \ NAURA Technology, \ Glodon, \ Ganfeng Lithium, \ Tianqi Lithium, \ Luxshare Precision, \ Rongsheng Petrochemical, \ 37 Interactive Entertainment, \ LB Group, \ Satellite Chemical, \ EVE Energy, \ Aier Eye Hospital, \ Zhifei Biological, \ Inovance Technology, \ Walvax Biotechnology, \ Sungrow Power Supply, \ Zhongji Innolight, \ Tigermed, \ Baotou Steel, \ China Northern Rare Earth, \ China Jushi, \ Hengli Petrochemical, \ Tongwei, \ Hundsun Technologies, \ Yonyou Network, \ Wingtech Technology, \ AECC Aviation Power, \ LONGi Green Energy, \ Aluminum Corporation of China (Chalco), \ China Tourism Group Duty Free, \ COSCO SHIPPING Holdings, \ CMOC Group & 44 \\
			3 & China Yangtze Power & 1 \\
			4 & Industrial and Commercial Bank of China (ICBC) & 1 \\
			5 & Sinopec, \ China Merchants Bank, \ Kweichow Moutai, \ PetroChina & 4 \\
			\hline
		\end{tabular}
		\label{tab:4-12}
	\end{table}
	
	In Table 4-12, comparing the second-order stochastic dominance relationships of the cluster centers reveals that $C_3 \equiv C_4 \equiv C_5$, and they stochastically dominate all other cluster centers in the second order. For the CSI 100 dataset, stocks in classes 3, 4, and 5 are selected to construct a suitable investment portfolio.
	
	\subsubsection{DSD-Hierarchical Clustering}
	
	Subsequently, stock hierarchical clustering is performed for risk-seeking investors, starting with first-order hierarchical clustering (FDSD-Hierarchical clustering), followed by second-order hierarchical clustering (SDSD-Hierarchical clustering). Both theoretical and experimental results indicate that the results of FASD-Hierarchical clustering and FDSD-Hierarchical clustering are the same. Therefore, for risk-seeking investors, this paper performs SDSD-Hierarchical clustering on the 75 NASDAQ constituent stocks and 76 CSI 100 constituent stocks. The optimal numbers of clusters are 8 and 10, respectively. The left chart in Figure 4-6 is the heatmap for NASDAQ, the right chart is for CSI 100. The figure shows good clustering effects for both datasets, with stocks clustered into classes with distinct differences in stochastic dominance coefficients.
	
	\begin{figure}[htbp]
		\subfloat[]{\includegraphics[width=0.45\textwidth]{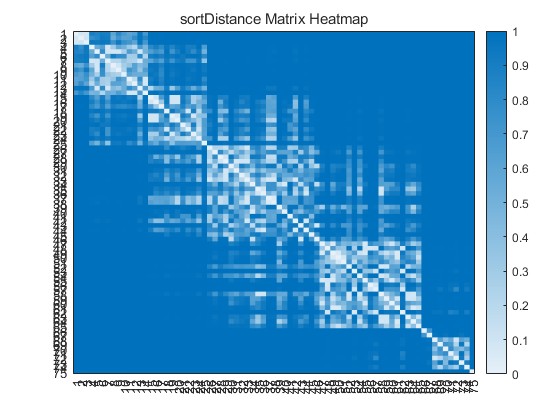}}
		\hfill 
		\subfloat[]{\includegraphics[width=0.45\textwidth]{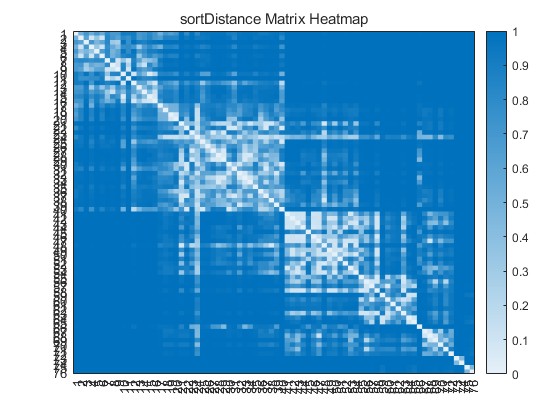}}
		\caption{SDSD-Hierarchical Stochastic Dominance Coefficient Heatmap}
		\label{fig:5}
	\end{figure}
	
	Table 4-13 details the situation where NASDAQ stocks are clustered into 8 classes. Comparing the second-order stochastic dominance relationships of the stock centers in the table, the centers of classes 4, 5, 6, and 7 stochastically dominate the centers of classes 1, 2, 3, and 8 in the second order. Therefore, the stocks in classes 4, 5, 6, and 7 of Table 4-13 are selected for investment choices.
	
	\begin{table}[htbp]
		\centering
		\caption{SDSD-Hierarchical Clustering Results (NASDAQ)}
		\setlength{\tabcolsep}{4pt} 
		\begin{tabular}{c p{10cm} c}
			\hline
			Cluster & Stock Names & Count \\
			\hline
			1 & American Electric Power,\ Mondelez International,\ Xcel Energy & 3 \\
			2 & Automatic Data Processing (ADP),\ Comcast,\ Costco Wholesale,\ Cisco Systems,\ Exelon,\ Honeywell International,\ Keurig Dr Pepper,\ Paychex,\ Roper Technologies,\ Starbucks,\ Verisk Analytics & 11 \\
			3 & Amgen,\ AstraZeneca (US ADR),\ Coca-Cola Europacific Partners,\ Cintas,\ Cognizant,\ Fastenal,\ Microsoft,\ O'Reilly Automotive,\ PACCAR,\ T-Mobile US,\ Texas Instruments & 11 \\
			4 & Apple,\ Adobe,\ Analog Devices,\ ANSYS,\ Cadence Design Systems,\ CDW Corp,\ Charter Communications,\ Copart,\ CoStar Group,\ Dollar Tree,\ Electronic Arts,\ Alphabet C (GOOG),\ Alphabet A (GOOGL),\ IDEXX Laboratories,\ Intel,\ Intuit,\ Monster Beverage,\ Old Dominion Freight Line,\ Ross Stores,\ Synopsys,\ Take-Two Interactive & 21 \\
			5 & Autodesk,\ Applied Materials,\ Amazon,\ ASML,\ Broadcom,\ Booking Holdings,\ Fortinet,\ Intuitive Surgical,\ KLA Corp,\ Lam Research,\ Lululemon Athletica,\ Marriott International,\ Microchip Technology,\ Meta Platforms,\ NXP Semiconductors,\ Palo Alto Networks,\ Qualcomm,\ Regeneron Pharmaceuticals,\ Vertex Pharmaceuticals & 19 \\
			6 & AMD,\ Tesla & 2 \\
			7 & DexCom,\ Diamondback Energy,\ MercadoLibre,\ Micron Technology,\ Netflix,\ NVIDIA,\ ON Semiconductor & 7 \\
			8 & PepsiCo & 1 \\
			\hline
		\end{tabular}
		\label{tab:4-13}
	\end{table}
	
	Table 4-14 shows the second-order clustering situation for CSI 100 stocks. Comparing the second-order stochastic dominance relationships of the stock centers in the table, it is found that the centers of classes 1 and 2 in Table 4-14 stochastically dominate other class centers in the second order. Therefore, the 16 stocks in classes 1 and 2 of the table are selected for investment choices.
	
	\begin{table}[htbp]
		\centering
		\caption{SDSD-Hierarchical Clustering Results (CSI 100)}
		\setlength{\tabcolsep}{4pt} 
		\begin{tabular}{c p{10cm} c}
			\hline
			Cluster & Stock Names & Count \\
			\hline
			1 & Unisplendour,\ Unigroup Guoxin Microelectronics,\ Tianqi Lithium,\ 37 Interactive Entertainment,\ Zhongji Innolight,\ Wingtech Technology & 6 \\
			2 & NAURA Technology,\ Ganfeng Lithium,\ EVE Energy,\ Zhifei Biological,\ Sungrow Power Supply,\ Tigermed,\ Tongwei,\ Hundsun Technologies,\ Yonyou Network,\ LONGi Green Energy & 10 \\
			3 & Focus Media,\ SF Holding,\ AECC Aviation Power,\ COSCO SHIPPING Holdings & 4 \\
			4 & ZTE Corporation,\ Eastern Shenghong,\ Changchun High-Tech,\ Qinghai Salt Lake Industry,\ TCL Zhonghuan Renewable Energy,\ iFLYTEK,\ Goertek,\ Oriental Yuhong,\ Glodon,\ Rongsheng Petrochemical,\ BYD,\ LB Group,\ Satellite Chemical,\ Inovance Technology,\ Walvax Biotechnology,\ China Northern Rare Earth,\ Hengli Petrochemical,\ Aluminum Corporation of China (Chalco),\ China Tourism Group Duty Free,\ CMOC Group & 20 \\
			5 & China Vanke,\ TCL Technology,\ BOE Technology,\ Hikvision,\ Aier Eye Hospital,\ Poly Developments,\ TBEA,\ China Jushi,\ Wanhua Chemical,\ NARI Technology,\ Hualu Hengsheng,\ Pien Tze Huang,\ Shandong Gold,\ Zijin Mining & 14 \\
			6 & Weichai Power,\ Gree Electric Appliances,\ Baoshan Iron \& Steel (Baosteel),\ SANY Heavy Industry,\ China Merchants Bank,\ Hengrui Medicine,\ Kweichow Moutai,\ Conch Cement,\ Haier Smart Home,\ China Shenhua Energy,\ Ping An Insurance & 11 \\
			7 & Luxshare Precision & 1 \\
			8 & Baotou Steel,\ CITIC Securities,\ China Unicom,\ China Railway Group,\ China State Construction Engineering (CSCEC),\ CRRC & 6 \\
			9 & China Yangtze Power,\ Industrial and Commercial Bank of China (ICBC) & 2 \\
			10 & Sinopec,\ PetroChina & 2 \\
			\hline
		\end{tabular}
		\label{tab:4-14}
	\end{table}
	
	\subsubsection{Model Performance Evaluation}
	
	The values of the classic Hierarchical clustering results are compared with those of the ASD-Hierarchical and DSD-Hierarchical clustering models. To ensure comparability, the number of clusters selected for the classic clustering models is kept consistent with that of the SD models.
	
	\begin{table}[htbp]
		\centering
		\caption{Comparison of SD-SC Values Between Classical and SD Clustering Models}
		\setlength{\tabcolsep}{12pt} 
		\begin{tabular}{ccccc}
			\toprule
			\multirow{2}{*}{} & \multicolumn{2}{c}{NASDAQ Dataset} & \multicolumn{2}{c}{CSI 100 Dataset} \\
			\cmidrule(lr){2-3} \cmidrule(lr){4-5}
			& SD & Classical & SD & Classical \\
			\midrule
			FASD(DSD)-HIE & 0.4301 & 0.2019 & & \\
			SASD-HIE      & 0.5460 & -0.5140 & 0.7167 & -0.0431 \\
			SDSD-HIE      & 0.3667 & -0.3195 & 0.3263 & -0.3097 \\
			\bottomrule
		\end{tabular}
		\label{tab:4-15}
	\end{table}
	
	\vspace{2em} 
	
	\begin{table}[htbp]
		\centering
		\caption{Comparison of SD-DBI Values Between Classical and SD Clustering Models}
		\setlength{\tabcolsep}{12pt}
		\begin{tabular}{ccccc}
			\toprule
			\multirow{2}{*}{} & \multicolumn{2}{c}{NASDAQ Dataset} & \multicolumn{2}{c}{CSI 100 Dataset} \\
			\cmidrule(lr){2-3} \cmidrule(lr){4-5}
			& SD & Classical & SD & Classical \\
			\midrule
			FASD(DSD)-HIE & 1.25 & 1.4509 & & \\
			SASD-HIE      & 2.2078 & 2.4571 & 2.2598 & 1.101 \\
			SDSD-HIE      & 1.6318 & 1.6602 & 8.4478 & 0.998 \\
			\bottomrule
		\end{tabular}
		\label{tab:4-16}
	\end{table}
	Table 4-15 presents the SD-SC values, and Table 4-16 presents the SD-DBI values. As seen in Table 4-15, the SD-SC values for all ASD-Hierarchical and DSD-Hierarchical clustering models across both datasets are higher than those of the classic clustering results. Table 4-16 shows that the SD-DBI values for the first-order and second-order ASD-Hierarchical and DSD-Hierarchical models on the NASDAQ dataset are all lower than the classic clustering results, verifying the quality of the clustering. However, the SD-DBI values for the CSI 100 dataset are all higher than the classic clustering results, which may be due to the weaker applicability of the SD-DBI index on this specific dataset.
	
	A comprehensive analysis combining both indices reveals that higher SD-SC coefficient values reflect better intra-class compactness and inter-class separation, while lower DBI values verify the rationality and effectiveness of the clustering results. Therefore, the index results demonstrate that the SD-Hierarchical clustering models proposed in this paper exhibit superior performance in stock clustering tasks targeting different investment preferences. Consequently, it can be concluded that the ASD-Hierarchical and DSD-Hierarchical clustering models possess significant advantages in applications regarding stock clustering based on investor preferences, providing more reliable references for investment decision-making.
	
	\subsection{SD-K-means Model Application}
	
	\subsubsection{$\alpha - \beta$ Plots}
	
		\begin{figure}[htbp] 
		\centering            
		\subfloat[NASDAQ SASD K-means $\alpha$-$\beta$]   
		{
			\label{fig:subfig5}\includegraphics[width=0.45\textwidth]{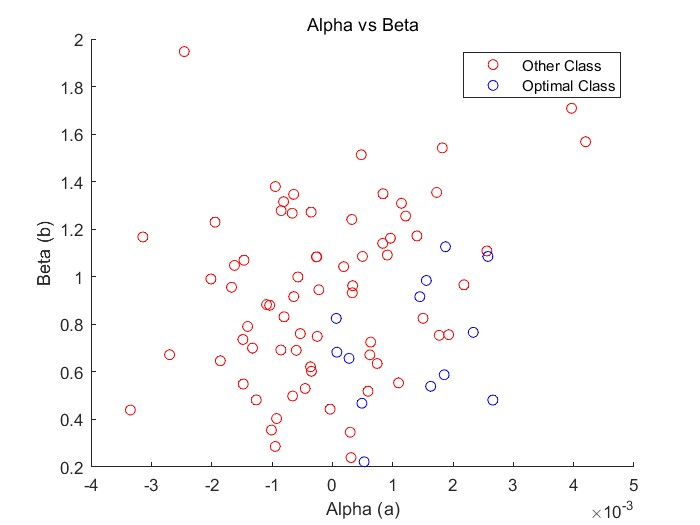}
		}
		\subfloat[NASDAQ  SDSD K-means $\alpha$-$\beta$]
		{
			\label{fig:subfig6}\includegraphics[width=0.45\textwidth]{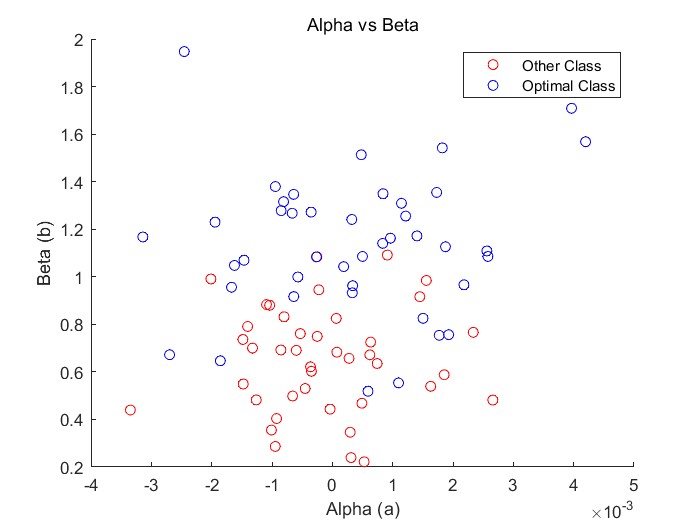}
		}
		\vspace{0.01cm} 
		\centering            
		\subfloat[CSI 100 SASD K-means $\alpha$-$\beta$]   
		{
			\label{fig:subfig7}\includegraphics[width=0.45\textwidth]{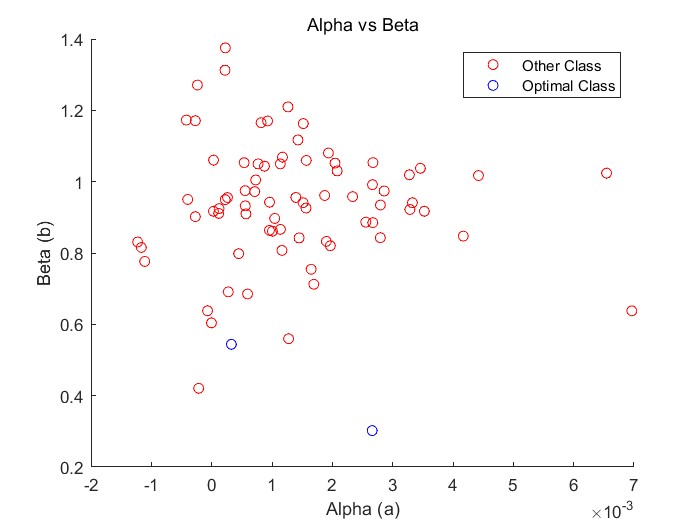}
		}
		\subfloat[CSI 100 SDSD K-means $\alpha$-$\beta$]
		{
			\label{fig:subfig8}\includegraphics[width=0.45\textwidth]{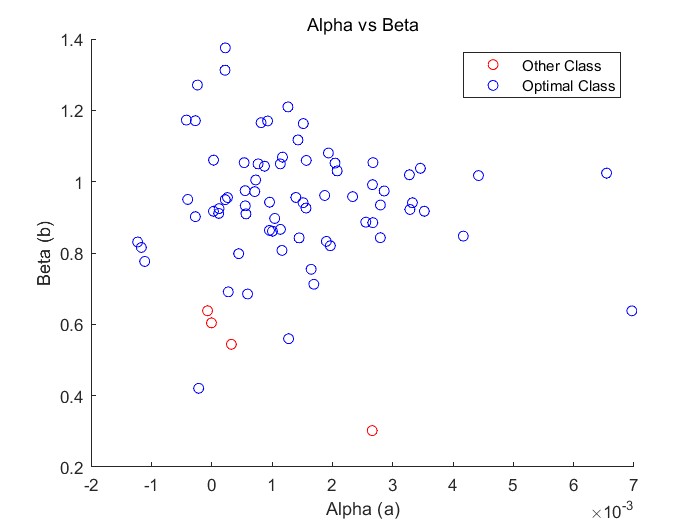}
		}
		\caption{$\alpha - \beta$ Plots for Different Datasets}  
		\label{Fig:7}           
	\end{figure}

	Figure 4-7 plots the excess return and risk volatility of NASDAQ constituents relative to the NASDAQ index average return, as well as CSI 100 constituents relative to the CSI 100 index average return. The horizontal axis $\alpha$ represents the excess return of a stock or portfolio relative to the market average return; the vertical axis $\beta$ represents the volatility of a stock or portfolio relative to the entire market, i.e., risk. Additionally, the relationship between stock return changes and overall market performance can be intuitively understood from the figure, providing wiser guidance for investors.
	
	From the perspective of creation potential, the difference between the optimal class of stocks selected and the unselected stocks is not significant, but there are significant differences in terms of market fluctuations: the $\beta$ values of stocks selected by the SASD-K-means algorithm are lower than those of other classes, locating these stocks in the lower half of the graph. Conversely, the $\beta$ values of stocks selected by the SDSD-K-means algorithm are opposite, located in the upper half. This aligns with the behavior of risk-averse individuals preferring low risk and low return, and risk-seekers preferring high risk and high return.
	
	\subsubsection{Portfolio Risk-Return Plots}
	
	\begin{figure}[htbp]
		\centering
		
		
		\begin{minipage}{0.48\textwidth}
			\centering
			\begin{minipage}{0.48\linewidth}
				\includegraphics[width=\linewidth]{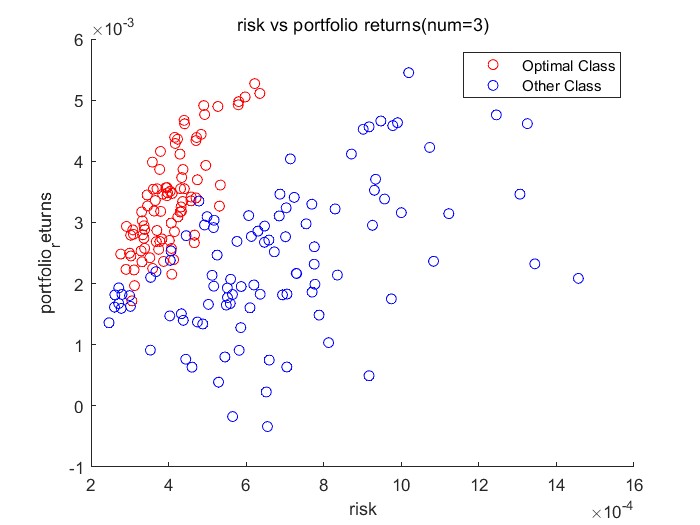}
			\end{minipage}
			\hfill
			\begin{minipage}{0.48\linewidth}
				\includegraphics[width=\linewidth]{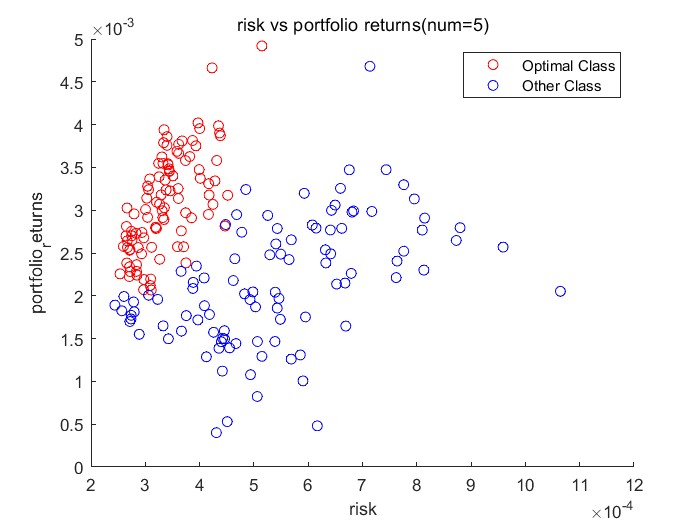}
			\end{minipage}
			
			\vspace{0.5em} 
			
			\begin{minipage}{0.48\linewidth}
				\includegraphics[width=\linewidth]{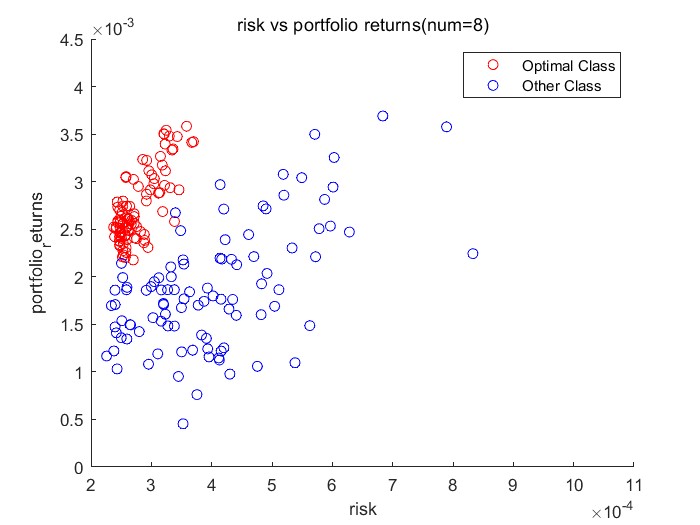}
			\end{minipage}
			\hfill
			\begin{minipage}{0.48\linewidth}
				\includegraphics[width=\linewidth]{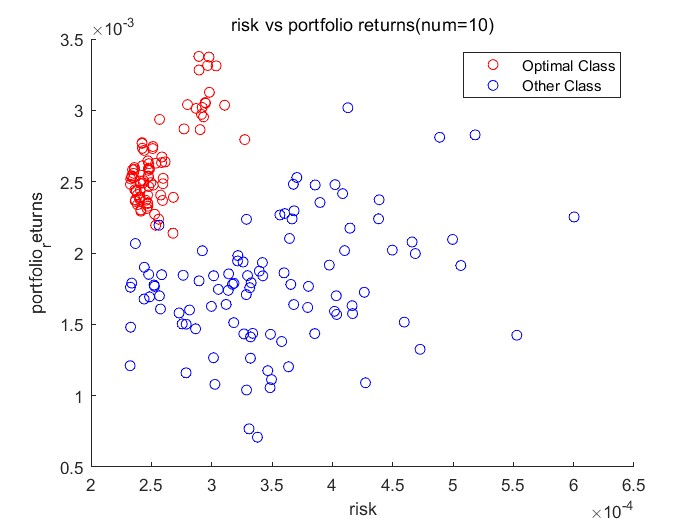}
			\end{minipage}
			
			\vspace{0.5em}
			\caption{NASDAQ ASD portfolio effect} 
			\label{fig:4-8}
		\end{minipage}
		\hfill
		\begin{minipage}{0.48\textwidth}
			\centering
			\begin{minipage}{0.48\linewidth}
				\includegraphics[width=\linewidth]{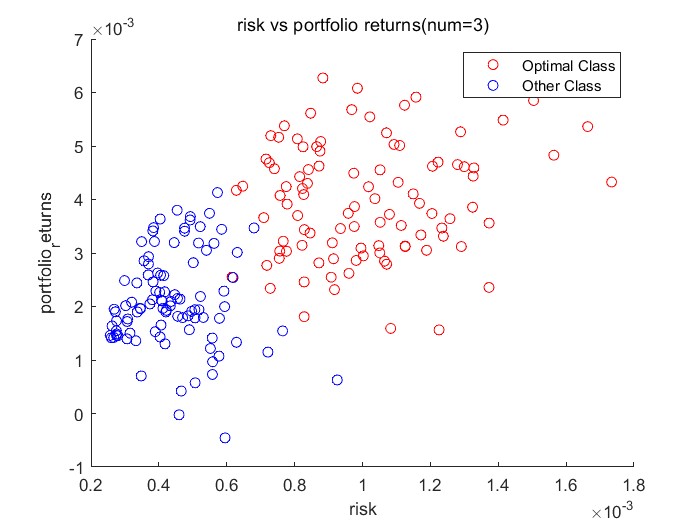}
			\end{minipage}
			\hfill
			\begin{minipage}{0.48\linewidth}
				\includegraphics[width=\linewidth]{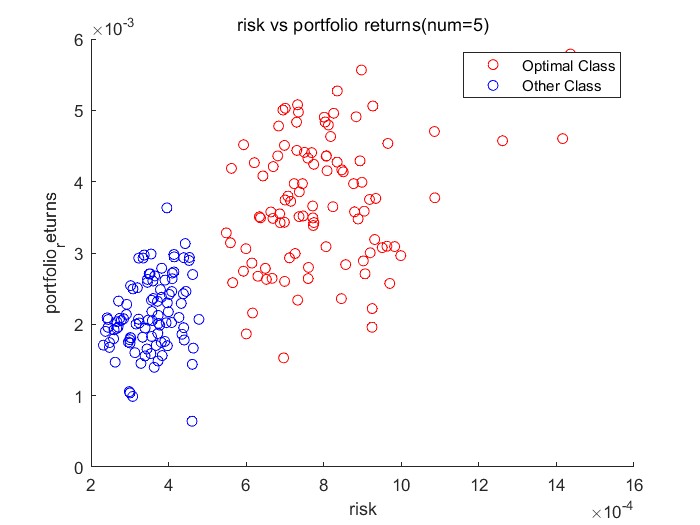}
			\end{minipage}
			
			\vspace{0.5em}
			
			\begin{minipage}{0.48\linewidth}
				\includegraphics[width=\linewidth]{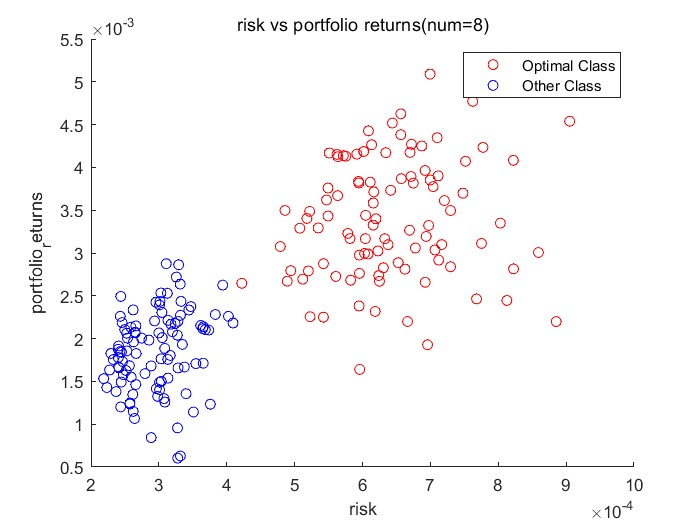}
			\end{minipage}
			\hfill
			\begin{minipage}{0.48\linewidth}
				\includegraphics[width=\linewidth]{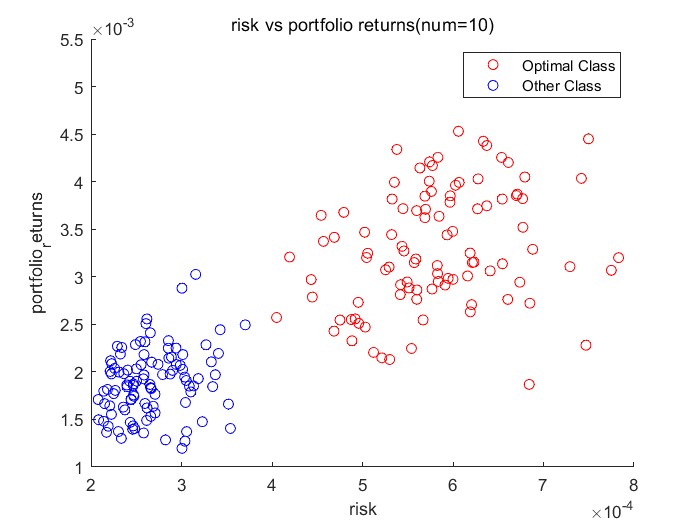}
			\end{minipage}
			
			\vspace{0.5em}
			\caption{NASDAQ DSD portfolio effect} 
			\label{fig:4-9}
		\end{minipage}
		
		\vspace{2em} 
		
		
		\begin{minipage}{0.48\textwidth}
			\centering
			\begin{minipage}{0.48\linewidth}
				\includegraphics[width=\linewidth]{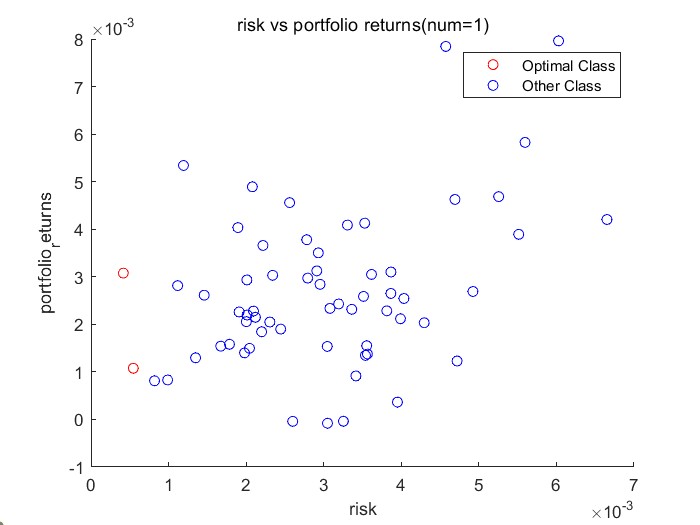}
			\end{minipage}
			\hfill
			\begin{minipage}{0.48\linewidth}
				\includegraphics[width=\linewidth]{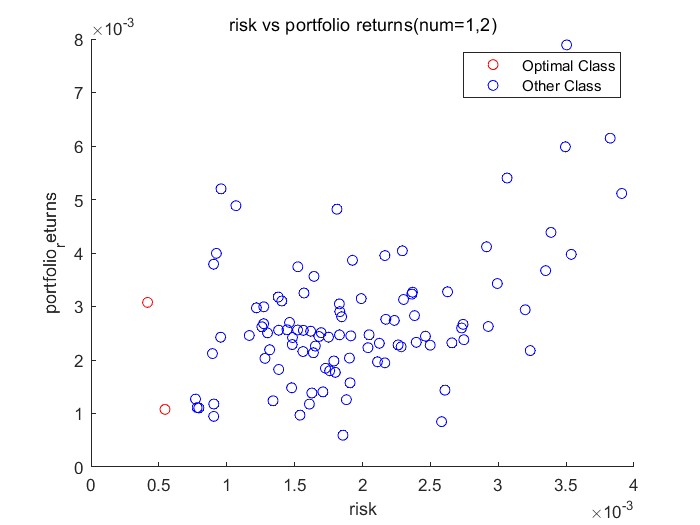}
			\end{minipage}
			
			\vspace{0.5em}
			
			\begin{minipage}{0.48\linewidth}
				\includegraphics[width=\linewidth]{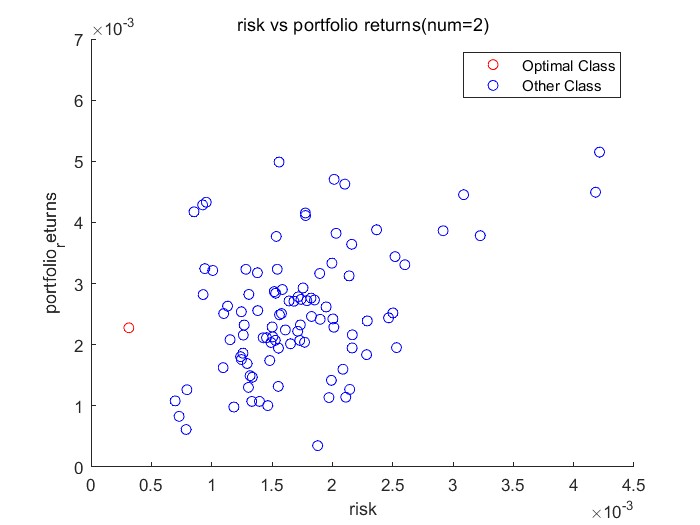}
			\end{minipage}
			\hfill
			\begin{minipage}{0.48\linewidth}
				\includegraphics[width=\linewidth]{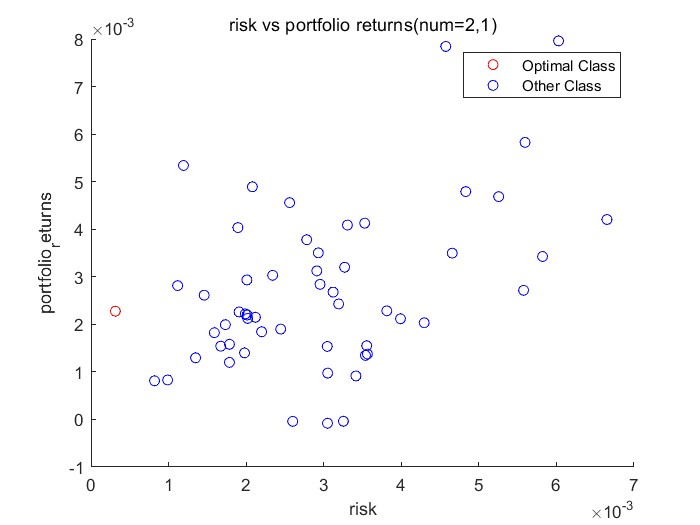}
			\end{minipage}
			
			\vspace{0.5em}
			\caption{CSI 100 ASD portfolio effect} 
			\label{fig:4-10}
		\end{minipage}
		\hfill
		\begin{minipage}{0.48\textwidth}
			\centering
			\begin{minipage}{0.48\linewidth}
				\includegraphics[width=\linewidth]{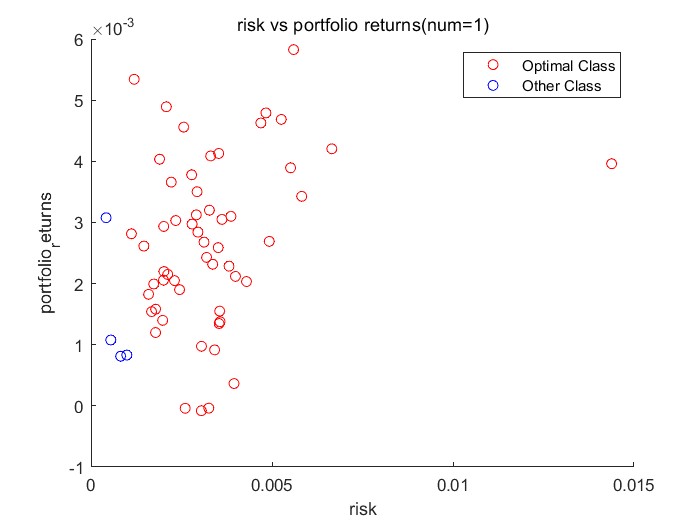}
			\end{minipage}
			\hfill
			\begin{minipage}{0.48\linewidth}
				\includegraphics[width=\linewidth]{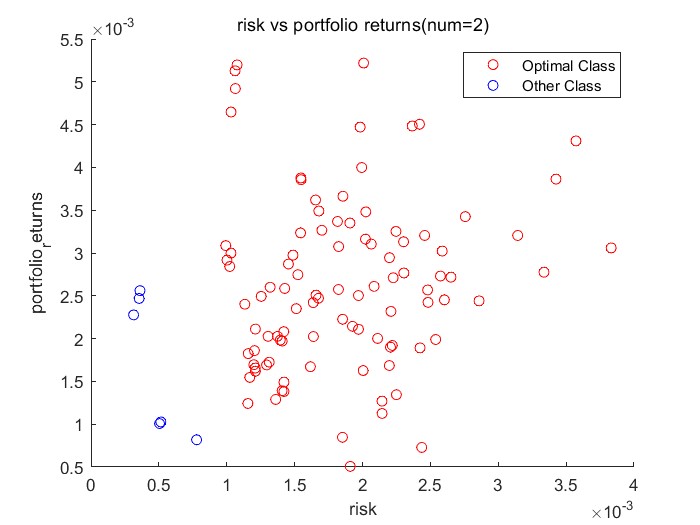}
			\end{minipage}
			
			\vspace{0.5em}
			
			\begin{minipage}{0.48\linewidth}
				\includegraphics[width=\linewidth]{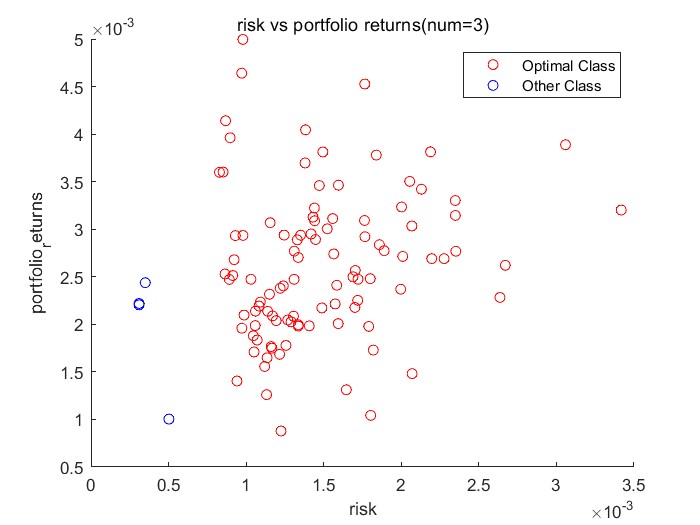}
			\end{minipage}
			\hfill
			\begin{minipage}{0.48\linewidth}
				\includegraphics[width=\linewidth]{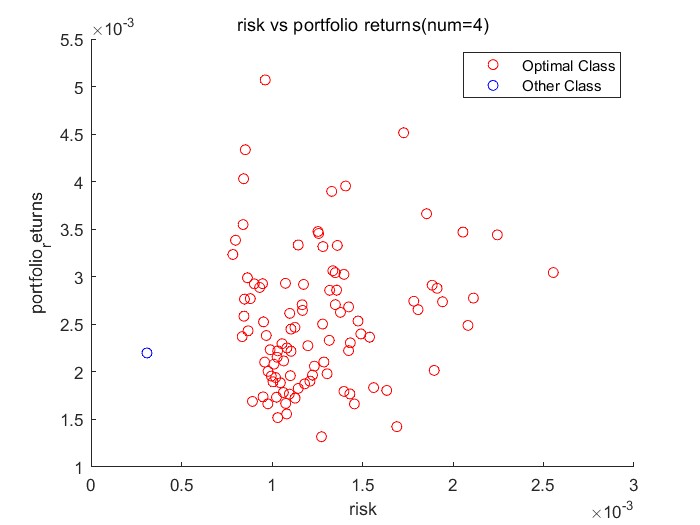}
			\end{minipage}
			
			\vspace{0.5em}
			\caption{CSI 100 DSD portfolio effect} 
			\label{fig:4-11}
		\end{minipage}
		
	\end{figure}
	
	Appropriate numbers of stocks are selected from the optimal class and the non-optimal class respectively to construct minimum variance portfolios, analyzing the risk and expected return of portfolios from different classes. The horizontal axis represents portfolio risk, and the vertical axis represents expected return. As seen in Figures 4-8 and 4-9, after clustering NASDAQ constituents, there are clear class relationships in the risk-expected return of portfolios constructed by drawing different numbers of stocks from different classes.
	
	For risk-averse investors, judging from the NASDAQ clustering results (Figure 4-8), portfolios constructed from stocks selected by the SASD-K-means algorithm possess low risk and high return, indicating significant algorithm effectiveness. Due to the smaller number of optimal stocks in CSI 100, no obvious cluster structure is visible (see Figure 4-10), but the selected stocks and constructed portfolios still possess low risk and low return characteristics compared to other stocks. When the number of stocks in the portfolio is 2, after 100 draws, the average return of the optimal class is 0.0023, lower than the non-optimal class's 0.0026; but simultaneously, the risk is 0.0004, also lower than the non-optimal class's 0.0032, aligning with risk-averse preferences.
	
	For risk-seeking investors, Figure 4-9 reveals that portfolios constructed from NASDAQ stocks selected by the SDSD-K-means algorithm possess high risk and high return, and there is an obvious cluster structure among portfolios; the more stocks in the portfolio, the clearer the structure. As seen in Figure 4-11, the effect for CSI 100 is relatively worse than NASDAQ due to the number of stocks, but still exhibits high risk and high return characteristics. For example, when the number of stocks is 2, after 100 draws, the average return of the optimal class is 0.0026, higher than the non-optimal class's 0.0017; but simultaneously, the risk is 0.00109, also higher than the non-optimal class's 0.00047, aligning with risk-seeking preferences.
	
	Through the above graphical analysis, it is found that: stochastic dominance clustering algorithms can effectively cluster numerous stocks into different classes based on investor preferences, and portfolios constructed from the selected optimal stocks also exhibit clear cluster structures. This verifies that the novel stochastic dominance clustering models proposed in this paper have significant application value in constructing minimum variance portfolios and can provide customized investment portfolio selection schemes for investor preferences. Furthermore, the new models are operable in risk control and asset allocation, allowing for asset allocation based on the risk levels investors can accept.
	
	\subsection{SD-Hierarchical Model Application}
	
	\subsubsection{$\alpha - \beta$ Plots}
	
		\begin{figure}[htbp] 
		\centering            
		\subfloat[NASDAQ SASD Hierarchical $\alpha$-$\beta$]   
		{
			\label{fig:subfig10}\includegraphics[width=0.45\textwidth]{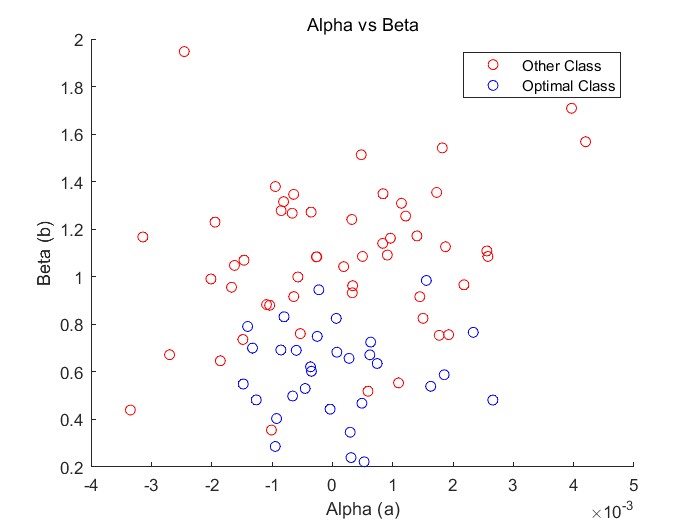}
		}
		\subfloat[NASDAQ SDSD Hierarchical $\alpha$-$\beta$]
		{
			\label{fig:subfig11}\includegraphics[width=0.45\textwidth]{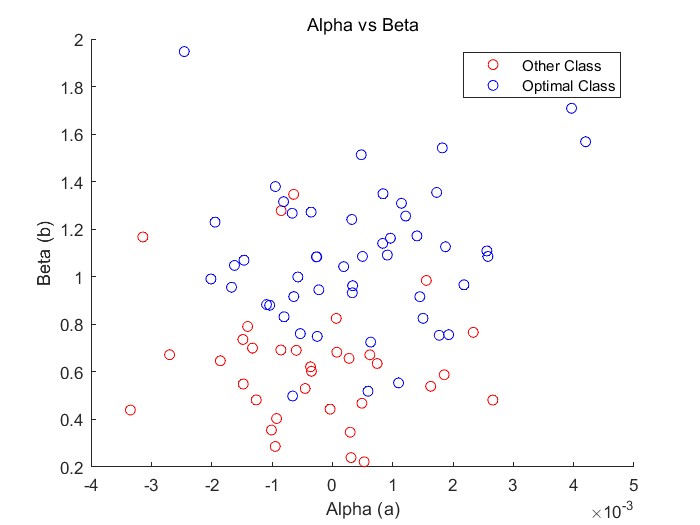}
		}
		\vspace{0.01cm} 
		\centering            
		\subfloat[CSI 100 SASD Hierarchical $\alpha$-$\beta$]   
		{
			\label{fig:subfig12}\includegraphics[width=0.45\textwidth]{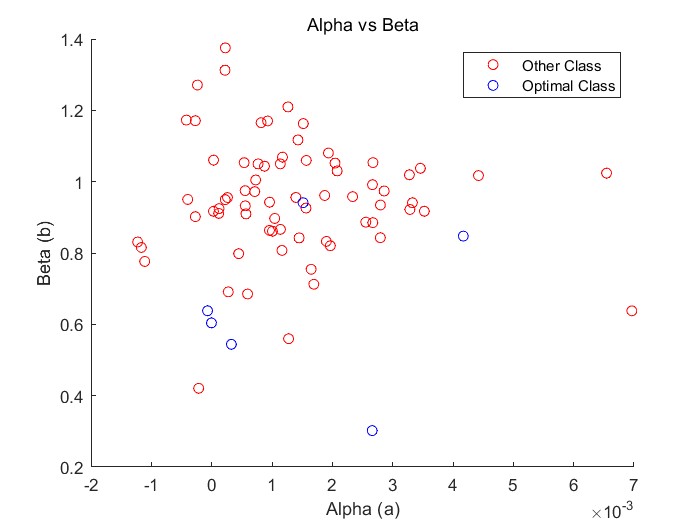}
		}
		\subfloat[CSI 100 SDSD Hierarchical $\alpha$-$\beta$]
		{
			\label{fig:subfig13}\includegraphics[width=0.45\textwidth]{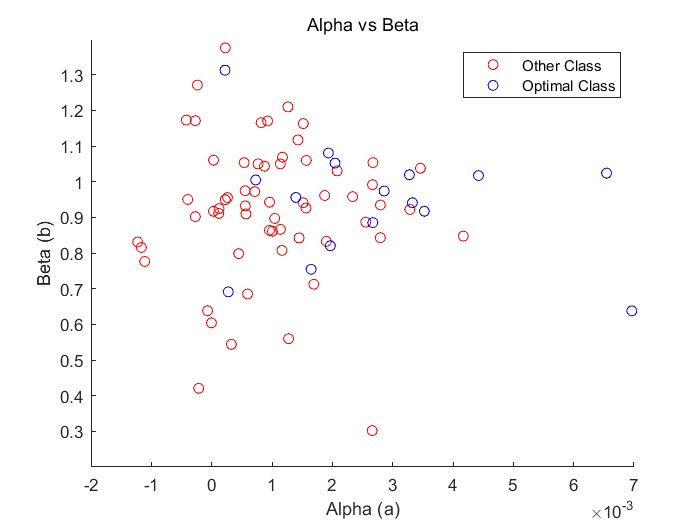}
		}
		\caption{$\alpha - \beta$ Plots for Different Datasets}   
		\label{Fig:12}           
	\end{figure}

	Figure 4-12 plots the excess return and risk volatility of NASDAQ constituents relative to the NASDAQ index average return, as well as CSI 100 constituents relative to the CSI 100 index average return. The horizontal axis $\alpha$ represents the excess return of a stock or investment portfolio relative to the market average return; the vertical axis $\beta$ represents the volatility of a stock or investment portfolio relative to the entire market, i.e., risk. From the perspective of creation potential, similar to the K-means clustering results, the difference between the optimal class of stocks selected and the unselected stocks is not significant; however, in terms of risk, there are differences in the risk of stocks selected by algorithms corresponding to different risk preferences. Additionally, the relationship between stock return changes and overall market performance can be intuitively understood from the figure, providing wiser guidance for investors to make decisions.
	
	The SASD-Hierarchical algorithm and SDSD-Hierarchical algorithm perform better on the NASDAQ dataset than on the CSI 100 constituent stocks. The $\alpha - \beta$ plot for NASDAQ stocks shows a clear cluster structure. The $\beta$ values of stocks selected by the SASD-Hierarchical algorithm are relatively low, while the $\beta$ values of stocks selected by the SDSD-Hierarchical algorithm are at a relatively high level. However, there is no obvious segmentation among CSI 100 stocks. The differences in the market structure and development potential of the US and Chinese stock markets are considered to lead to the difference in results on the $\alpha - \beta$ plots. Next, this paper selects an appropriate number of stocks from different classes to construct minimum variance portfolios, compares the risk and expected return of portfolios from different classes, and discusses the performance of the proposed algorithms in another application scenario.

	\subsubsection{Portfolio Risk-Return Plots}
	
	\begin{figure}[htbp]
		\centering
		
		
		\begin{minipage}{0.48\textwidth}
			\centering
			\begin{minipage}{0.48\linewidth}
				\includegraphics[width=\linewidth]{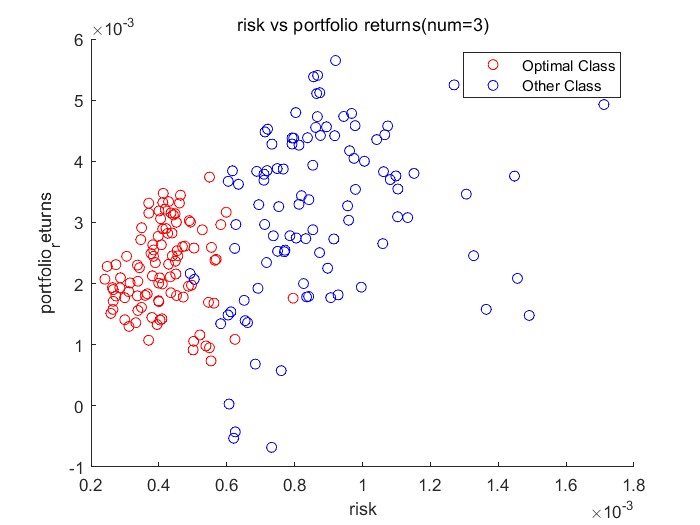}
			\end{minipage}
			\hfill
			\begin{minipage}{0.48\linewidth}
				\includegraphics[width=\linewidth]{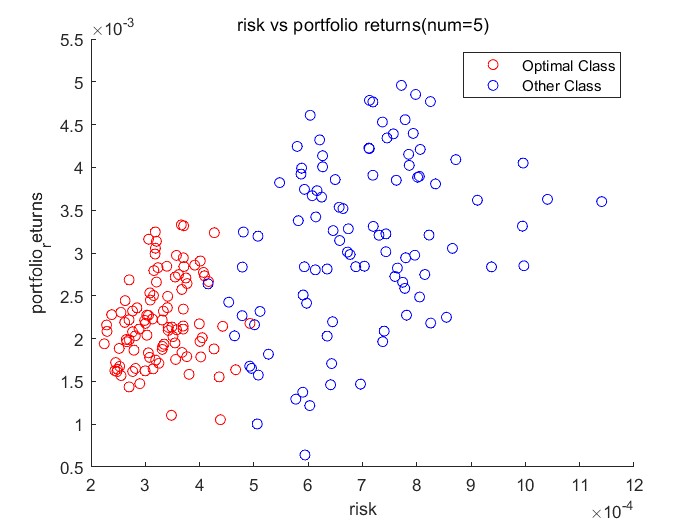}
			\end{minipage}
			
			\vspace{0.5em} 
			
			\begin{minipage}{0.48\linewidth}
				\includegraphics[width=\linewidth]{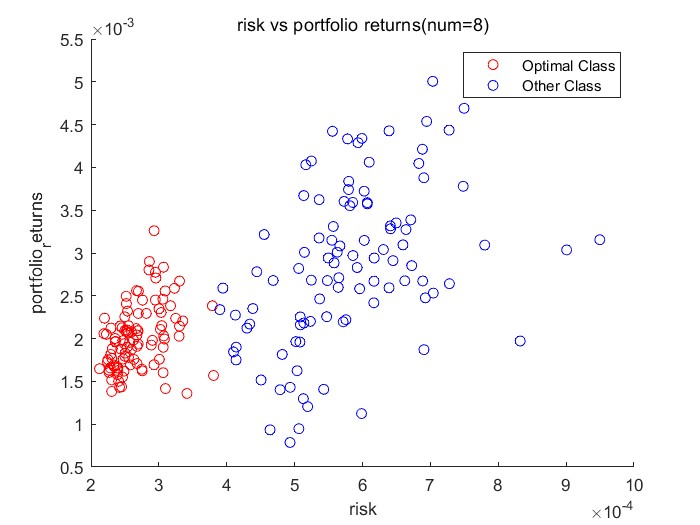}
			\end{minipage}
			\hfill
			\begin{minipage}{0.48\linewidth}
				\includegraphics[width=\linewidth]{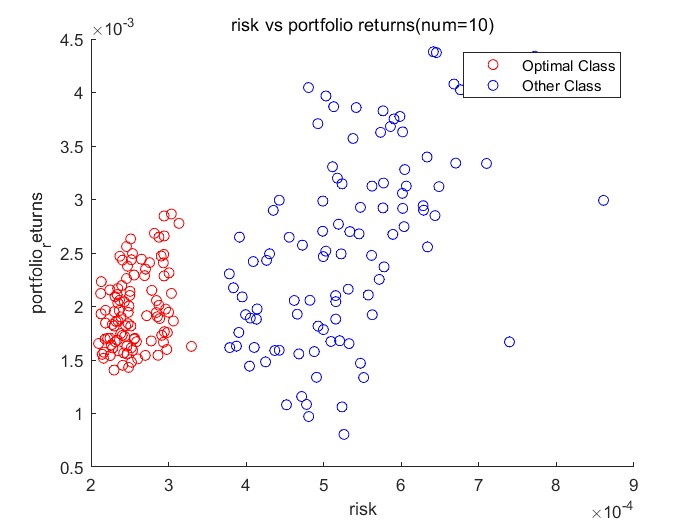}
			\end{minipage}
			
			\vspace{0.5em}
			\caption{NASDAQ ASD portfolio effect} 
			\label{fig:4-13}
		\end{minipage}
		\hfill
		\begin{minipage}{0.48\textwidth}
			\centering
			\begin{minipage}{0.48\linewidth}
				\includegraphics[width=\linewidth]{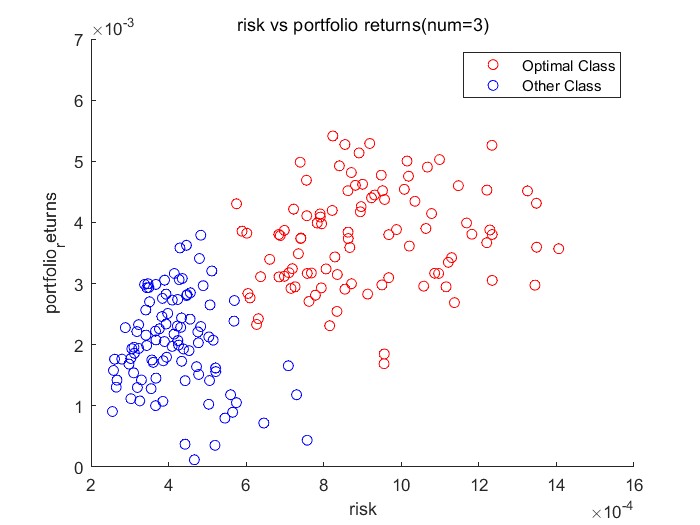}
			\end{minipage}
			\hfill
			\begin{minipage}{0.48\linewidth}
				\includegraphics[width=\linewidth]{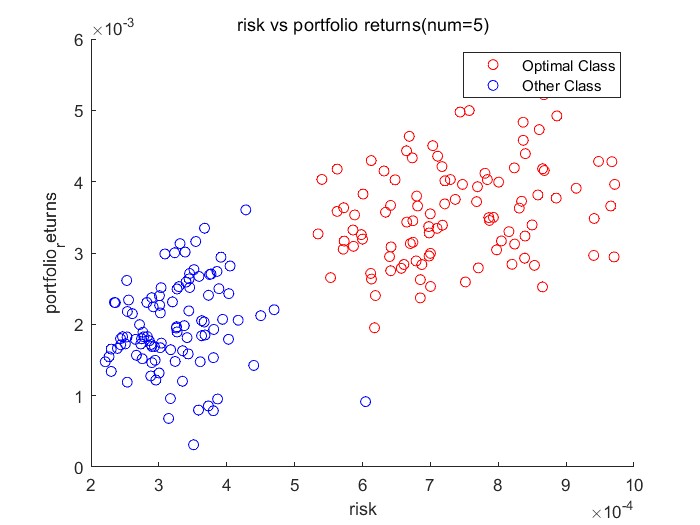}
			\end{minipage}
			
			\vspace{0.5em}
			
			\begin{minipage}{0.48\linewidth}
				\includegraphics[width=\linewidth]{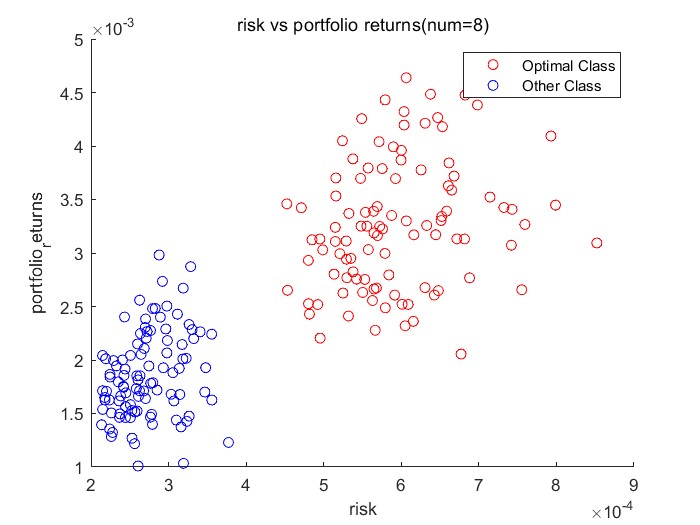}
			\end{minipage}
			\hfill
			\begin{minipage}{0.48\linewidth}
				\includegraphics[width=\linewidth]{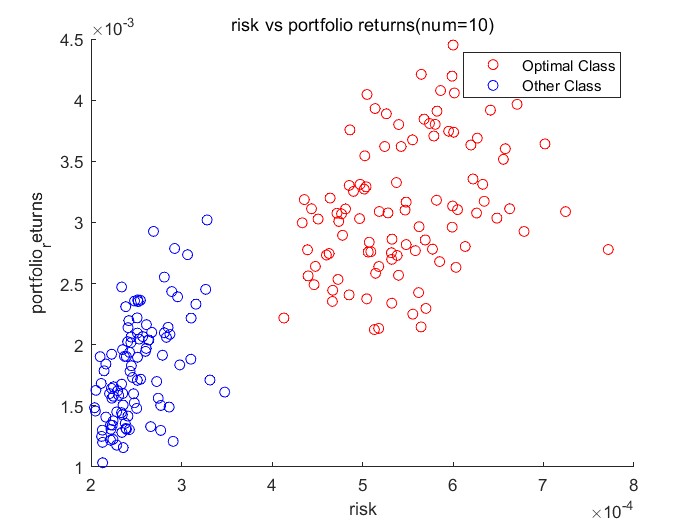}
			\end{minipage}
			
			\vspace{0.5em}
			\caption{NASDAQ DSD portfolio effect} 
			\label{fig:4-14}
		\end{minipage}
		
		\vspace{2em} 
		
		
		\begin{minipage}{0.48\textwidth}
			\centering
			\begin{minipage}{0.48\linewidth}
				\includegraphics[width=\linewidth]{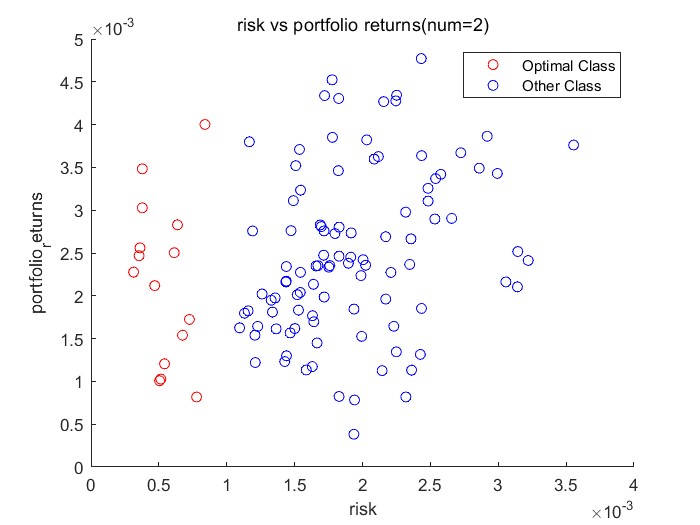}
			\end{minipage}
			\hfill
			\begin{minipage}{0.48\linewidth}
				\includegraphics[width=\linewidth]{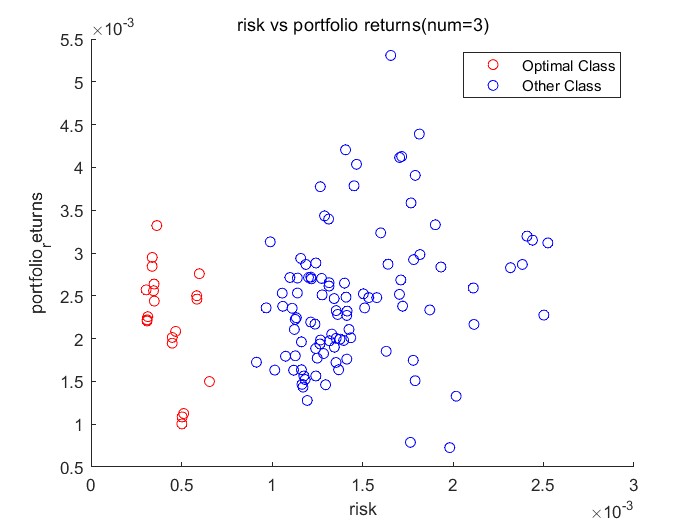}
			\end{minipage}
			
			\vspace{0.5em}
			
			\begin{minipage}{0.48\linewidth}
				\includegraphics[width=\linewidth]{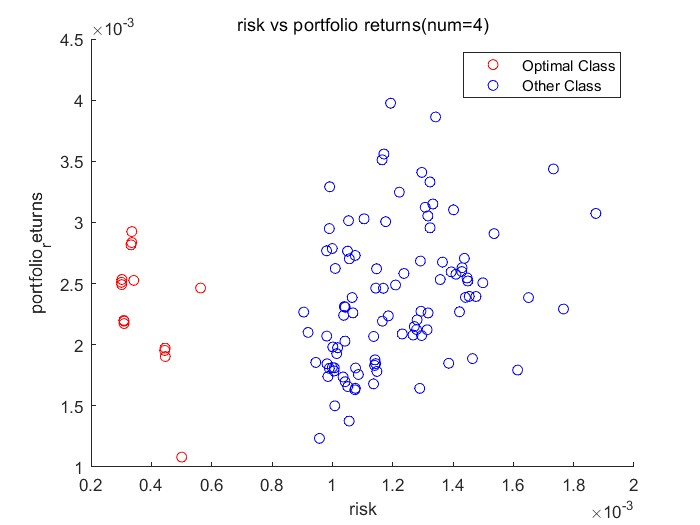}
			\end{minipage}
			\hfill
			\begin{minipage}{0.48\linewidth}
				\includegraphics[width=\linewidth]{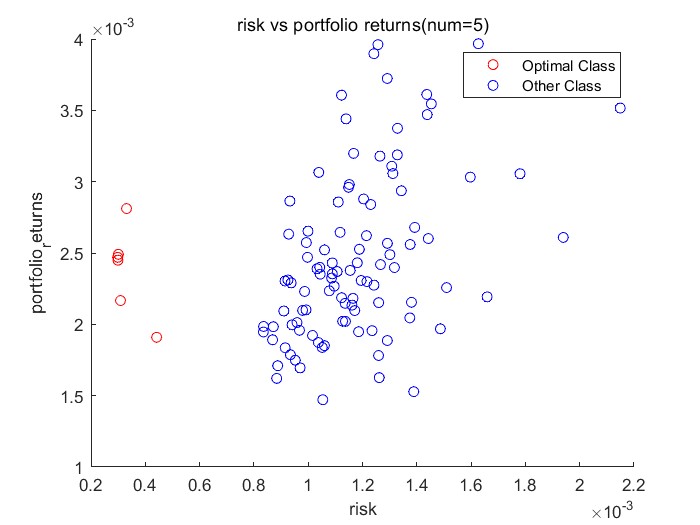}
			\end{minipage}
			
			\vspace{0.5em}
			\caption{CSI 100 ASD portfolio effect} 
			\label{fig:4-15}
		\end{minipage}
		\hfill
		\begin{minipage}{0.48\textwidth}
			\centering
			\begin{minipage}{0.48\linewidth}
				\includegraphics[width=\linewidth]{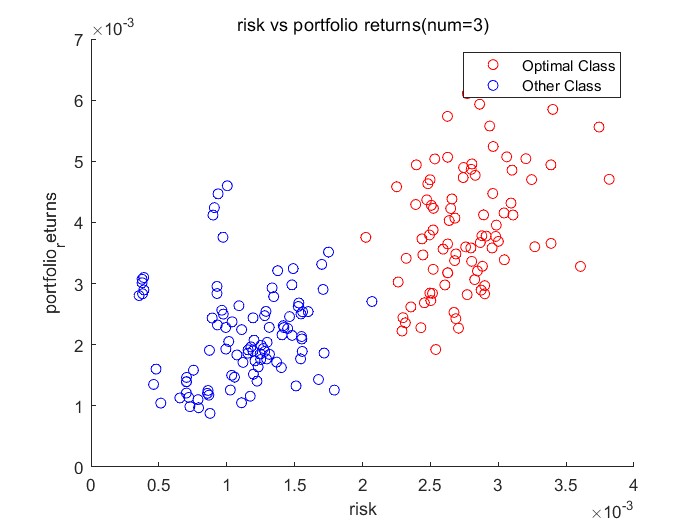}
			\end{minipage}
			\hfill
			\begin{minipage}{0.48\linewidth}
				\includegraphics[width=\linewidth]{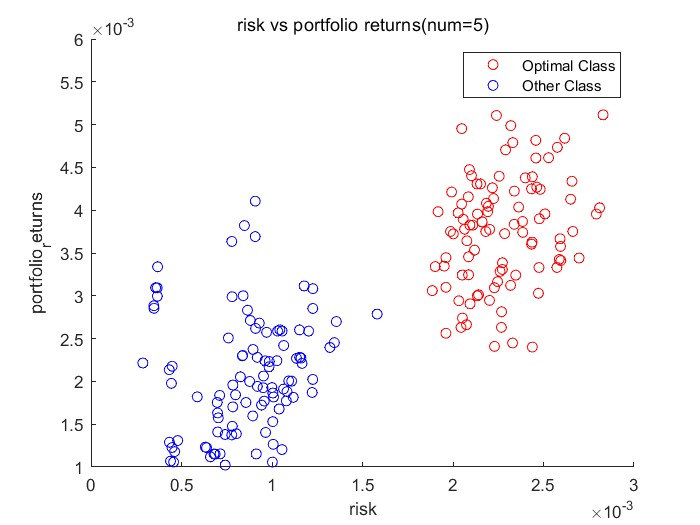}
			\end{minipage}
			
			\vspace{0.5em}
			
			\begin{minipage}{0.48\linewidth}
				\includegraphics[width=\linewidth]{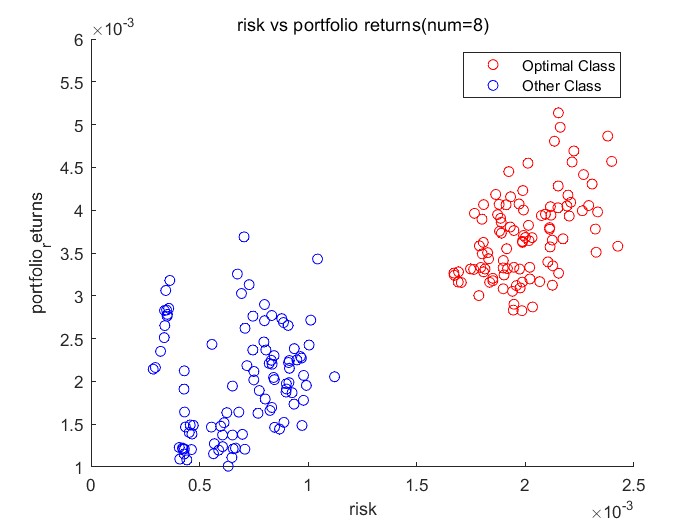}
			\end{minipage}
			\hfill
			\begin{minipage}{0.48\linewidth}
				\includegraphics[width=\linewidth]{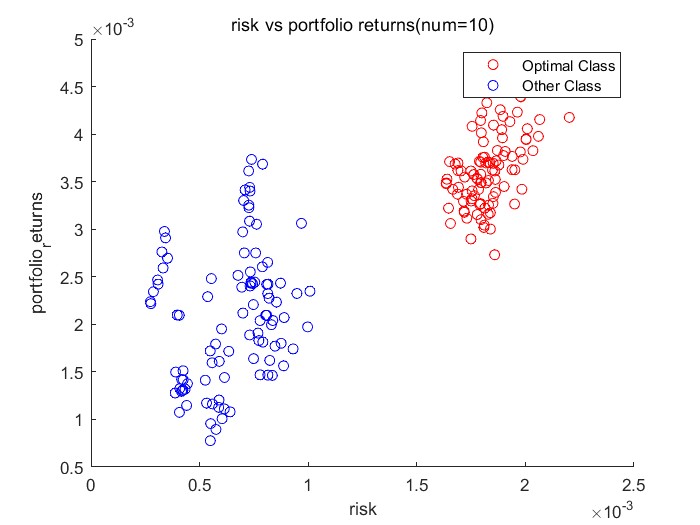}
			\end{minipage}
			
			\vspace{0.5em}
			\caption{CSI 100 DSD portfolio effect} 
			\label{fig:4-16}
		\end{minipage}
		
	\end{figure}
	
	In the figures, the horizontal axis represents the risk of the investment portfolio, and the vertical axis represents the expected return of the portfolio. Figures 4-13 and 4-15 are the risk-return plots of investment portfolios constructed from stocks selected by risk-averse investors. The portfolios constructed from the optimal class of stocks have relatively lower returns compared to portfolios constructed from other classes of stocks, but simultaneously, the risk is also low. Figures 4-14 and 4-16 are the risk-return plots of investment portfolios constructed from stocks selected by the SDSD-Hierarchical algorithm for the two datasets, both exhibiting high risk and high return characteristics. The investment portfolios constructed from stocks obtained by these two hierarchical clustering methods show clear cluster structures between classes. As the number of investment portfolios increases, the cluster structure becomes more obvious, and the risk-expected return of stocks within the same class is compact and similar.
	
	The risk-return distribution plots above indicate: stochastic dominance clustering algorithms can effectively cluster numerous stocks into different classes based on investor preferences, and the investment portfolios constructed from the selected optimal stocks also exhibit clear cluster structures. This verifies that the novel stochastic dominance clustering models proposed in this paper have significant application value in constructing minimum variance portfolios and can provide customized investment portfolio selection schemes for investor preferences. Furthermore, the new models are operable in risk control and asset allocation, allowing for reasonable asset allocation based on the risk levels investors can accept.
	
	A comprehensive comparison of the clustering performance of ASD-K-means, ASD-Hierarchical, DSD-K-means, and DSD-Hierarchical algorithms across different datasets reveals: judging from the distribution plots of investment portfolios, compared to the DSD-K-means algorithm, the clustering effect of the DSD-Hierarchical algorithm is more significant, and the inter-class separation is greater, meaning the Hierarchical algorithm is better able to cluster stocks into classes with significant stochastic dominance relationships. In addition, it can also be found that the performance of NASDAQ constituent stock data on stochastic dominance clustering models is superior to that of CSI 100 constituent stock data.
	
	\section{Conclusion and Outlook}
	
	This paper proposed corresponding SD-K-means clustering algorithms and SD-Hierarchical clustering algorithms based on the first-order, second-order, and third-order stochastic dominance relationships for risk-averse and risk-seeking investors, respectively. Using the clustering algorithms and the SD-SC coefficient to select the optimal number of clusters, the NASDAQ index constituents and CSI 100 index constituents were clustered. After clustering, the class center was taken as the representative of each class to compare the stochastic dominance relationships between cluster centers, and the class where the center stochastically dominating other class centers is located was selected for subsequent investment or portfolio construction.
	
	Since using the first-order and second-order stochastic dominance relationships of stocks can already divide stocks into classes with obvious structures, this paper used first-order and second-order clustering algorithms. If the first-order and second-order clustering algorithms cannot well divide stocks into classes with obvious stochastic dominance relationships, one can further consider third-order or higher-order K-means clustering and Hierarchical clustering algorithms to cluster stocks.
	
	When constructing the SD-SC coefficient and SD-DBI index to evaluate the clustering results, it was found that for the NASDAQ dataset, the final clustering results of the clustering models in this paper had higher SD-SC coefficient values and lower DBI values compared to classic K-means clustering and Hierarchical clustering models. This reflects better intra-class compactness and inter-class separation, confirming the rationality of the clustering results of the new algorithms proposed in this paper. The SD-SC coefficient values for the CSI 100 index dataset were also higher than those of the classic K-means clustering and Hierarchical clustering models, but the SD-DBI values were all higher than the classic clustering results, which is considered to be due to the weaker applicability of the SD-DBI index on this dataset.
	
	In addition, the application effects of the SD clustering models on single index models and the construction of global minimum variance portfolios were explored, which also verified that the algorithms proposed in this paper can well provide customized asset allocation schemes for investors with different risk preferences, possessing great practical application value. A comprehensive comparative analysis of the effects shows: compared to the SD-K-means clustering algorithm, the SD-Hierarchical algorithm is better able to cluster stocks into classes with significant stochastic dominance relationships, and the clustering effect is superior. Furthermore, the performance of NASDAQ constituent stock data on stochastic dominance clustering models is superior to that of CSI 100 constituent stock data.
	
	The analysis in this paper also found that the algorithms proposed herein can divide stocks into two categories based on their stochastic dominance relationships: for risk-averse investors, the selected stocks have the characteristics of low risk and low return; for risk-seeking investors, the selected stocks provide higher returns and higher risks. The expected returns and returns of investment portfolios constructed from different categories of stocks also show different cluster structures. This indicates that the algorithms proposed in this paper can well provide assistance to investors with different preferences when making investment decisions.
	
	Although the SD-Hierarchical clustering algorithm and SD-K-means clustering algorithm proposed in this paper have achieved relatively good results in the clustering analysis of NASDAQ 100 index and CSI 100 constituents, there are still some deficiencies. Firstly, the sample scope is limited: currently, empirical research is only conducted on specific markets (such as US stocks and partial A-share constituents) and has not yet been expanded to the entire stock market, which may affect the universality of the model. Secondly, currently, only the SD-SC coefficient and SD-DBI index have been constructed to evaluate the clustering results, and a more comprehensive investor preference-oriented evaluation system has not yet been constructed.
	
	Based on the results and deficiencies of this study, future research can be deepened from the following aspects: First, consider expanding the research scope to conduct empirical analysis on the entire stock market. In addition, this paper will further optimize the SD-K-means clustering algorithm and modify the distance matrices of other classic clustering algorithms to improve more algorithms, compare the performance of different algorithms, and conduct a more comprehensive and scientific evaluation of the effectiveness and applicability of various clustering methods under different market conditions on a larger scale. Furthermore, consider constructing other stochastic dominance clustering evaluation indices and deeply analyzing the applicability issue of the SD-DBI index on the CSI 100 dataset, in order to comprehensively verify that the clustering models in this paper have significant advantages in investor preference-oriented stock clustering applications, providing more reliable references for risk-averse and risk-seeking investors when making investment decisions.
	\bibliographystyle{unsrtnat}
	\bibliography{kylSDKC.bib}

@mastersthesis {Yang2019,
author = { Yang Xiaojun },
title = {Research on K-means clustering algorithm and its application in stock investment},
school = {Chongqing University},
year = {2019}
}

@article{Korzeniewski2018, 
  title={Efficient stock portfolio construction by means of clustering},
  author={Korzeniewski, Jerzy},
  journal={Acta Universitatis Lodziensis. Folia Oeconomica},
  volume={1},
  number={333},
  pages={85--92},
  year={2018}
}

@inproceedings{Huang2010,
  title={Clustering Effect of Style Based on Pearson Correlation: Example of SSE50 Sample Stocks},
  author={Huang, xue Fei and Zhao, Xin and Li, Cheng},
  booktitle={2010 International Conference on Internet Technology and Applications},
  pages={1--4},
  year={2010},
  organization={IEEE}
}

@inproceedings{Lu2020,
  title={A New Cluster Validity Index for Stock Clustering Based on Efficient Frontier},
  author={Lu, Yahui and Li, Minghao and Tang, Xiaochu and Wang, Hui},
  booktitle={2020 5th IEEE International Conference on Big Data Analytics (ICBDA)},
  pages={193--197},
  year={2020},
  organization={IEEE}
}

@article{Lúcio2022,
  title={COVID-19 and stock market volatility: A clustering approach for S\&P 500 industry indices},
  author={Lúcio, Francisco and Caiado, Jorge},
  journal={Finance Research Letters},
  volume={49},
  pages={103141},
  year={2022},
  publisher={Elsevier}
}

@article{Wu2022,
  title={Construction of stock portfolios based on k-means clustering of continuous trend features},
  author={Wu, Dingming and Wang, Xiaolong and Wu, Shaocong},
  journal={Knowledge-Based Systems},
  volume={252},
  pages={109358},
  year={2022},
  publisher={Elsevier}
}

@article{Siregar2024, 
  title={Implementation of K-means clustering algorithm for the Indonesian stock exchange},
  author={Siregar, Bakti and Yosia, Yosia},
  journal={Jurnal Sisfotek Global},
  volume={14},
  number={1},
  pages={49--56},
  year={2024}
}

@article{Guy2014,
  title={Upside and downside beta portfolio construction: A different approach to risk measurement and portfolio construction},
  author={Guy, CFA and others},
  journal={Available at SSRN 2612235},
  year={2014}
}

@article{Boloș2025,
  title={K-Means Clustering for Portfolio Optimization: Symmetry in Risk--Return Tradeoff, Liquidity, Profitability, and Solvency},
  author={Boloș, Marcel-Ioan and Rusu, Ștefan and Leordeanu, Marius and Sab{\u{a}}u-Popa, Claudia Diana and Perțicaș, Diana Claudia and Crișan, Mihai-Ioan},
  journal={Symmetry},
  volume={17},
  number={6},
  pages={847},
  year={2025},
  publisher={MDPI}
}

@incollection{McFadden1989,
  title = {Testing for Stochastic Dominance},
  booktitle = {Studies in the Economics of Uncertainty: In Honor of Josef Hadar},
  author = {McFadden, Daniel},
  editor = {Fomby, Thomas B. and Seo, Tae Kun},
  year = {1989},
  pages = {113--134},
  publisher = {Springer New York},
  address = {New York, NY},
  isbn = {978-1-4613-8924-8 978-1-4613-8922-4}
}

@article{Klecan1991,
  title = {A Robust Test for Stochastic Dominance},
  author = {Klecan, Lindsey and McFadden, Raymond and McFadden, Daniel},
  year = {1991},
  journal = {working paper, Department of Economics, MIT}
}

@article{Barrett2003,
  title = {Consistent Tests for Stochastic Dominance},
  author = {Barrett, Garry F. and Donald, Stephen G.},
  year = {2003},
  journal = {Econometrica},
  volume = {71},
  number = {1},
  pages = {71--104}
}

@article{Linton2005,
  title = {Consistent Testing for Stochastic Dominance under General Sampling Schemes},
  author = {Linton, Oliver and Maasoumi, Esfandiar and Whang, Yoon-Jae},
  year = {2005},
  journal = {The Review of Economic Studies},
  volume = {72},
  number = {3},
  pages = {735--765}
}

@article{McAleer2006,
  title = {Stochastic Dominance Test for Risk Seekers: An Application to Oil Spot and Futures Markets},
  shorttitle = {Stochastic Dominance Test for Risk Seekers},
  author = {McAleer, Michael and Lean, Hooi Hooi and Wong, Wing-Keung},
  year = {2006},
  journal = {SSRN Electronic Journal}
}

@book{Sriboonchitta2009,
  title = {Stochastic Dominance and Applications to Finance, Risk and Economics},
  author = {Sriboonchitta, Songsak and Wong, Wing-Keung and Dhompongsa, s and Nguyen, Hung},
  year = {2009},
  publisher = {{Chapman and Hall/CRC}},
  isbn = {978-0-429-13814-0}
}

@article{Bai2015,
  title = {Stochastic Dominance Statistics for Risk Averters and Risk Seekers: An Analysis of Stock Preferences for USA and China},
  shorttitle = {Stochastic Dominance Statistics for Risk Averters and Risk Seekers},
  author = {Bai, Zhidong and Li, Hua and McAleer, Michael and Wong, Wing-Keung},
  year = {2015},
  journal = {Quantitative Finance},
  volume = {15},
  number = {5},
  pages = {889--900}
}

@article{Levy1985,
  title = {Upper and Lower Bounds of Put and Call Option Value: Stochastic Dominance Approach},
  shorttitle = {Upper and Lower Bounds of Put and Call Option Value},
  author = {Levy, Haim},
  year = {1985},
  journal = {The Journal of Finance},
  volume = {40},
  number = {4},
  pages = {1197--1217}
}

@article{Post2003,
  title = {Empirical Tests for Stochastic Dominance Efficiency},
  author = {Post, Thierry},
  year = {2003},
  journal = {The Journal of Finance},
  volume = {58},
  number = {5},
  pages = {1905--1931}
}

@article{Wong2007,
  title = {Stochastic Dominance and Mean--Variance Measures of Profit and Loss for Business Planning and Investment},
  author = {Wong, Wing-Keung},
  year = {2007},
  journal = {European Journal of Operational Research},
  volume = {182},
  number = {2},
  pages = {829--843}
}

@phdthesis{Li2013,
  title = {Test Statistics for Prospect and Markowitz Stochastic Dominances with Applications},
  author = {Li, Hua},
  year = {2013},
  journal = {(A) Mathematics/ Physics/ Mechanics/ Astronomy},
  collaborator = {Bai, Zhidong},
  school = {Northeast Normal University}
}

@article{Li1999,
  title = {Extension of Stochastic Dominance Theory to Random Variables},
  author = {Li, Chi-Kwong and Wong, Wing-Keung},
  year = {1999},
  journal = {RAIRO - Operations Research - Recherche Op{\'e}rationnelle},
  volume = {33},
  number = {4},
  pages = {509--524}
}

@article{Davidson2000,
  title = {Statistical Inference for Stochastic Dominance and for the Measurement of Poverty and Inequality},
  author = {Davidson, Russell and Duclos, Jean-Yves},
  year = {2000},
  journal = {Econometrica},
  volume = {68},
  number = {6},
  pages = {1435--1464}
}
\end{sloppypar}
\end{document}